\documentclass[runningheads]{llncs}
\usepackage{graphicx}

\usepackage{amsmath}
\usepackage{amssymb}

\usepackage{multirow}
\usepackage{booktabs}
\usepackage{subcaption}
\usepackage{float}
\usepackage{enumitem}
\usepackage[ruled]{algorithm2e}
\usepackage[utf8]{inputenc}
\usepackage[T1]{fontenc}
\usepackage{textcomp}
\newcommand{\cut}[1]{}

\newcommand{\nameS}{S-MLDG}
\newcommand{\nameFS}{FO-\nameS}
\newcommand{\nameFFS}{FFO-\nameS}
\newcommand{\hundo}{S-Undo-Bias}
\newcommand{\keypoint}[1]{\vspace{0.1cm}\noindent\textbf{#1}\quad}
 
\begin{document}
\title{Sequential Learning for Domain Generalization}
\titlerunning{Sequential Learning for Domain Generalization}
\author{Da Li$^{1}$\thanks{Work done while DL was at SketchX}, Yongxin Yang$^2$, Yi-Zhe Song$^2$ and Timothy Hospedales$^{1,2,3}$}
\authorrunning{Li et al.}
\institute{Samsung AI Centre, Cambridge, UK \and
SketchX, CVSSP, University of Surrey, UK
% \url{http://www.springer.com/gp/computer-science/lncs} 
\and
The University of Edinburgh, UK\\
\email{dali.academic@gmail.com, \{yongxin.yang,y.song\}@surrey.ac.uk, t.hospedales@ed.ac.uk}
}
\maketitle              % typeset the header of the contribution
\begin{abstract}
In this paper we propose a sequential learning framework for Domain Generalization (DG), the problem of training a model that is robust to domain shift by design. Various DG approaches have been proposed with different motivating intuitions, but they typically optimize for a single step of domain generalization -- training on one set of domains and generalizing to one other. Our sequential learning is inspired by the idea lifelong learning, where accumulated experience means that learning the $n^{th}$ thing becomes easier than the $1^{st}$ thing. In DG this means encountering a sequence of domains and at each step training to maximise performance on the next domain. The performance at domain $n$ then depends on the previous $n-1$ learning problems. Thus backpropagating through the sequence means optimizing performance not just for the next domain, but all following domains. Training on all such sequences of domains provides dramatically more `practice' for a base DG learner compared to existing approaches, thus improving performance on a true testing domain. This strategy can be instantiated for different base DG algorithms, but we focus on its application to the recently proposed Meta-Learning Domain generalization (MLDG). We show that for MLDG it leads to a simple to implement and fast algorithm that provides consistent performance improvement on a variety of DG benchmarks. 

\end{abstract}
\section{Introduction}
\label{sec:introduction}

% \epigraph{\emph{``Practice makes perfect"}}{John Adams}
% The performances of the machine learning systems usually drop dramatically when they meet the domain with statistical difference than the training domain. In the machine learning community, it is known as the domain shift between training and testing domains.

Contemporary machine learning algorithms provide excellent performance when training and testing data are drawn from the same underlying distribution. However, it is often impossible to guarantee prior collection of training data that is representative of the environment in which a model will be deployed, and the resulting train-test domain shift leads to significant degradation in performance. The long studied area of \emph{Domain Adaptation} (DA) aims to alleviate this by adapting models to the testing domain \cite{shai2006nipsdomainadaptation,tzeng2014deep,long2015learning,ganin2016dann,mslongnips2016,bousmalis2016domain}. Meanwhile, the recently topical area of \emph{Domain Generalization} (DG) aims to build or train models that are designed for increased robustness to such domain-shift without requiring adaptation~\cite{muandet2013domaingeneralization,ghifary2015domain,da2017dg,Li2018MLDG,mmdaaecvpr2018,shiv2018dg,NIPS2018_metareg}.

A variety of DG methods have been proposed based on different intuitions. To learn a domain-agnostic feature representation, some of these require specific base learner architectures \cite{muandet2013domaingeneralization,ghifary2015domain,ECCV12_Khosla}. Others are model-agnostic modifications to the training procedure of any base learner, for example by via data augmentation \cite{shiv2018dg,riccardo_nips18}. Meta-learning (a.k.a learning to learn) has a long history  \cite{Schmidhuber95onlearning,1991bengiolearningtolearn}, with primary application to accelerating learning of new tasks \cite{ravi2016optimization,pmlr-v80-wei18a}. Recently, some researchers proposed  meta-learning based methods for DG \cite{Li2018MLDG,NIPS2018_metareg}. Different from previous  DG methods, these are designed around explicitly mimicking train-test domain-shift during model training, to develop improved robustness to domain-shift at testing.  Such meta-learning has an analogy to human learning, where a human's experience of context change provides the opportunity to develop strategies that are more agnostic to context (domain). If a human discovers that their existing problem-solving strategy fails in a new context, they can try to update their strategy to be more context independent, so that next time they arrive in a new context they are more likely to succeed immediately. 

While effective, recent meta-DG methods \cite{Li2018MLDG,NIPS2018_metareg} provide a `single-step' of DG meta-learning: training on one set of training domains to optimize performance on a  disjoint set of `validation' domains. However, in human lifelong learning, such learning does not happen once, but sequentially in a continual learning manner. Taking this perspective in algorithm design, one learning update from domain $n$ to $n+1$ should have the opportunity to affect the performance on every subsequent domain encountered, $n+2$ onwards. Such continual learning provides the opportunity for much more feedback to each learning update. In backpropagation, the update at domain $n\to n+1$ can be informed by its downstream impact on all subsequent updates for all subsequent domains. In this way we can generate more unique episodes for meta-learning, which has improved performance in the more common few-shot applications of meta-learning \cite{vinyals2016oneShot,pmlr-v80-wei18a,lopezpaz2017GEM}. Specifically, in approaches that use a single-pass on all source domains~\cite{ECCV12_Khosla,motiian2017CCSA,ganin2016dann}, DG models are trained once for a single objective. Approaches doing one-step meta-learning  \cite{Li2018MLDG,NIPS2018_metareg} by rotating through meta-train and meta-test (validation) domain splits of $N$ source domains train DG with $N$ distinct domain-shift episodes. Meanwhile within our sequential learning DG framework, by further simulating all possible sequential learning domain sequences, we train with $N!$ distinct domain-shift episodes. This greater diversity of domain-shift training experience enables better generalization to a final true testing domain. 

Our proposed framework can be instantiated for multiple base DG algorithms without modifying their underlying design. We focus on its instantiation for a recent architecture-agonstic meta-learning based method MLDG \cite{Li2018MLDG}, but also show that it can be applied to a  traditional architecture based method Undo Bias \cite{ECCV12_Khosla}. In the case of MLDG, we show  our sequential-learning generalization \nameS{}, leads to a simple to implement and fast to train meta-learning algorithm that is architecture agnostic and consistently improves performance on a variety of DG benchmarks. This is achieved via a first-order approximation to the full \nameS{}, which leads to a shortest-path descent method analogous to Reptile \cite{DBLP:journals/corr/abs-1803-02999} in few-shot learning.

We summarise our contributions as follows:
\begin{itemize}[noitemsep]%,nolistsep]
    % it is a general sequential learning framework?
    \item We propose a sequential learning framework for DG that can be applied to different base DG methods. We show that it can be instantiated for at least two different base DG methods, the architecture focused Undo-Bias \cite{ECCV12_Khosla}, and the architecture agnostic meta-learning algorithm MLDG \cite{Li2018MLDG}. 
    %, one is a recently proposed meta-learning based DG method -- MLDG \cite{Li2018MLDG} and another is the traditional underlying domain learning DG method -- Undo Bias \cite{ECCV12_Khosla}. Our hierarchical training does not require any change on the structure of the base-DG methods. By integrating the base method to our lifelong learning framework it improves the performance of base method directly.
    \item Our framework improves training by increasing the diversity of unique DG episodes constructed for training the base learner, and enabling future changes in continual-learning performance changes to back-propagate to earlier domain updates. 
    %    Our lifelong learning framework is flexible to provide a diversity of tasks for training the model. This can be regarded as task augmentation for making base methods `practice' at the task of interest under sufficient times. Then, the trained models have better chance of performing well at real testing.
    \item We provide an analysis of the proposed \nameS{}, to understand the difference in optimization to the base MLDG algorithm, and to derive a fast first-order approximation \nameFFS{}. This algorithm is simple to implement and fast to run, while performing comparably to \nameS{}. 
    %. Meanwhile, we use a shortest path descent method, which is similar to `Reptile'~\cite{DBLP:journals/corr/abs-1803-02999}, as a faster first-order approximation for \nameS{}. This approximator leads to simple implementation and saves run-time computation, but with comparable performance to \nameS{}. Furthermore, we theoretically analyze that this efficient approximator works in a way similar to \nameS{}.
    \item The resulting \nameS{}  and \nameFFS{} algorithms provide state of the art performance on three different DG benchmarks.
\end{itemize}

\section{Related Work}

% \keypoint{Multi-Domain Learning (MDL)} Usually the domain generalization tasks begin with a few source domains. Learning from the multiple source domains is a topical problem \cite{rebuff2017mdlvisdecath, Rebuffi18}. But, MDL per se focuses on building the efficient learning across the different source domains and benefiting the performances within source domains. This is different from domain generalization, which cares the testing performances out of source domains.

\keypoint{Domain Adaptation (DA)} Domain adaptation has received great attention from researchers in the past decade \cite{shai2006nipsdomainadaptation,tzeng2014deep,long2015learning,ganin2015unsupervised,mslongnips2016,bousmalis2016domain,pmlr-v70-saito17a,saito2018MCD}. Different from domain generalization, domain adaptation assumes that unlabeled target domain data is accessible at training. Various methods have been proposed to tackle domain-shift by reducing discrepancy between source and target domain features. Representative approaches include aligning domains by minimizing distribution shift as measured by MMD \cite{tzeng2014deep,long2015learning}, or performing adversarial training to ensure that in the learned representation space the domains cannot be distinguished  \cite{ganin2015unsupervised,saito2018MCD}, or learning generative models for cross-domain image synthesis \cite{NIPS2016_6544,hoffman2018cycada}. However, data may not be available for the target domain, or it may not be possible to adapt the base model, thus requiring Domain Generalization.

\keypoint{Domain Generalization (DG)} A diversity of DG methods have been proposed in recent  years \cite{muandet2013domaingeneralization,ECCV12_Khosla,Xu2014lre,ghifary2015domain,motiian2017CCSA,da2017dg,Li2018MLDG,shiv2018dg,mmdaaecvpr2018,NIPS2018_metareg,riccardo_nips18}. These are commonly categorized according to their motivating inductive bias, or their architectural assumptions. Common motivating intuitions include feature learning methods \cite{ghifary2015domain,muandet2013domaingeneralization,mmdaaecvpr2018} that aim to learn a representation that generates domain invariant features; data augmentation-based methods that aim to improve robustness by synthesizing novel data that better covers the space of domain variability compared to the original source domains \cite{riccardo_nips18,shiv2018dg}; and fusion methods that aim to perform well on test domains by recombining classifiers trained on diverse source domains \cite{Xu2014lre,Massimiliano2018ICIP}. Meanwhile in terms of architecture, some methods impose constraints on the specific base classifier architecture to be used \cite{muandet2013domaingeneralization,ECCV12_Khosla,ghifary2015domain,da2017dg,Xu2014lre}, while others provide an architecture agnostic DG training strategy \cite{NIPS2018_metareg,shiv2018dg,riccardo_nips18}.

Most of the above methods train a single set of source tasks for a DG objective. Recent meta-learning methods use the set of known source domains to simulate train-test domain-shift and optimize to improve robustness to domain-shift. For example, via gradient alignment \cite{Li2018MLDG} or meta-optimizing a robust regularizer for the base model \cite{NIPS2018_metareg}. Our sequential learning framework aims to simulate continual learning over a sequence of domains, and furthermore averages over many such sequences. This provides a greater diversity of distinct domain-shift experiences to learn from, and stronger feedback in the form of the impact of a parameter change not just on the next validation domain, but its subsequent impact on all domains in the continual learning sequence. We show that our framework can be instantiated primarily for the meta-learning method MLDG \cite{Li2018MLDG}, but also for the classic architecture-specific method Undo-Bias \cite{ECCV12_Khosla}, and its recently proposed deep extension  \cite{da2017dg}.

\keypoint{Meta-Learning} Meta-Learning (learning to learn) has a long history \cite{Schmidhuber95onlearning}. It has recently become widely used in few-shot learning \cite{andrychowicz2016learning,ravi2016optimization,finn2017model,metanetworks2017} applications. A common meta-optimization strategy is to split training tasks into meta-train and meta-test (validation) task sets, and meta-optimization aims to improve the ability to learn quickly on meta-test tasks given the knowledge in meta-train tasks. This is achieved through various routes, by learning a more general feature embedding \cite{vinyals2016oneShot,sung2018learning}, learning a more efficient optimizer \cite{andrychowicz2016learning,ravi2016optimization}, or even simply learning an effective initial condition for optimization \cite{finn2017model,DBLP:journals/corr/abs-1803-02999}. Several gradient-based meta-learners induce higher-order gradients that increase computational cost, for example MAML \cite{finn2017model}. This inspired other studies to develop first order approximations for faster meta-learning; such as Reptile \cite{DBLP:journals/corr/abs-1803-02999} that accelerates MAML.  While all these methods meta-optimize for fast adaptation to new tasks, we aim to optimize for domain-generalization: training a model such that it performs well on a novel domain with no opportunity for adaptation. We take inspiration from Reptile \cite{DBLP:journals/corr/abs-1803-02999} to develop a fast implementation of our proposed \nameS{}.

\keypoint{Lifelong Learning} Our sequential learning is inspired by the vision of lifelong learning (LLL) \cite{pentina2015lllnoniid,ruvolo2013ella,schmidhuber1997inductiveBias,lopezpaz2017GEM}
. LLL methods focus on how to accelerate learning of new tasks given a series of sequentially learned previous tasks (and often how to avoid forgetting old tasks). We leverage the idea of optimizing for future performance in a sequence. But different to prior methods: (i) we focus on optimizing for domain invariance, rather than optimizing for speed of learning a new task, and (ii) we back-propagate through the entire sequence of domains so that every update step in the sequence is driven by improving the final domain invariance of the base model. It is important to note that while most lifelong and continual learning studies are oriented around designing a method that is \emph{deployed} in a lifelong learning setting, we address a standard problem setting with a fixed set of source and target (testing) domains. We aim to use sequential training within our given source domains to learn a more robust model that generalizes better to the true testing domain. To this end, since different potential learning sequences affect the outcome in lifelong learning \cite{lampert2015curriculum}, we aim to generate the most unique learning experiences to drive training by simulating all possible sequences through our source domains and optimizing for their expected outcome. 

\section{Domain Generalization Background}\label{sec:bg}
In the domain generalization problem setting, a learner receives $N$ labelled domains (datasets) $\mathcal{D}=[\mathcal{D}_1, \mathcal{D}_2, \cdots, \mathcal{D}_N]$ where $\mathcal{D}_i = (X_i, y_i)$, and aims to produce a model that works for a different \emph{unseen} domain $\mathcal{D}_*$ at testing. \cut{It assumes that the domains are homogeneous (containing corresponding input channels and label spaces).} We first introduce a simple baseline for DG .

\keypoint{Aggregation Baseline} A simple baseline for DG is to aggregate all domains' data and train a single model on $\mathcal{D}_{\text{agg}}=\mathcal{D}_1\cup\mathcal{D}_2\cup\dots\cup\mathcal{D}_N$. Although not always compared, this obvious baseline often outperforms earlier published DG methods when applied with deep learning \cite{da2017dg}.

\keypoint{Base Methods} Our sequential learning framework can be applied to generalize MLDG \cite{Li2018MLDG} and shallow \cite{ECCV12_Khosla} or deep \cite{da2017dg} Undo-Bias. Due to space constraints, we focus on the application to MLDG, and leave application to Undo-Bias to Appendix~\ref{appendix:sec:undo}. 

\subsection{Meta-Learning Domain Generalization}\label{sec:mldg}
In contrast to many DG methods \cite{ECCV12_Khosla,da2017dg,mmdaaecvpr2018,shiv2018dg}, which require special designs of model architectures, \emph{Meta-Learning Domain Generalization} (MLDG) \cite{Li2018MLDG} proposes an optimization method to achieve DG that is agnostic to base learner architecture. The idea is to mimic (during training) the cross-domain training and testing encountered in the DG setting -- by way of meta-training and meta-testing steps. 

In each iteration of training it randomly selects one domain $\mathcal{D}_{k}, k \in [1, N]$ and uses it as the meta-test domain, i.e. $\mathcal{D}_{\text{mtst}}\leftarrow \mathcal{D}_{k}$ (here $\mathcal{D}_{\text{mtst}}$ can be seen as a kind of \emph{virtual} test domain), and aggregates the remaining to construct the meta-train domain, i.e., $\mathcal{D}_{\text{mtrn}}\leftarrow\underset{i\neq k}{\cup} \mathcal{D}_i$.

Following the intuition that meta-test will be used to evaluate the effect of the model optimization on meta-train at each iteration, MLDG aims to optimize both the loss on meta-train  $\mathcal{L}_{1}= \mathcal{L}(\mathcal{D}_{\text{mtrn}}, \theta)$, and loss on meta-test after updating on meta-train $\mathcal{L}_{2}=\mathcal{L}(\mathcal{D}_{\text{mtst}}, \theta - \alpha \cdot{\nabla_{\theta}\mathcal{L}_{1})}$ by one gradient descent step  $\alpha \cdot \nabla_{\theta}\mathcal{L}_{1}$ with step size $\alpha$, where $\mathcal{L}(.)$ is the cross-entropy loss here. Overall this leads to optimization of

\small
\begin{equation}
\label{eq-mldg}
\underset{\theta}{\operatorname{argmin}}~\mathcal{L}_1(\mathcal{D}_{\text{mtrn}}, \theta) + \beta\mathcal{L}_2(\mathcal{D}_{\text{mtst}}, \theta - \alpha \nabla_\theta\mathcal{L}_1).
\end{equation}
\normalsize

\noindent After training, the base model with parameters $\theta$ will be used for true unseen test domain.

%Besides MLDG a traditional DG method \emph{Undo Bias} is another base DG method we will extend.
%The background of \emph{Undo Bias} is introduced in Appendix \ref{appendix:sec:undo}.

\begin{figure}[t]
\centering
\includegraphics[width=1.0\linewidth]{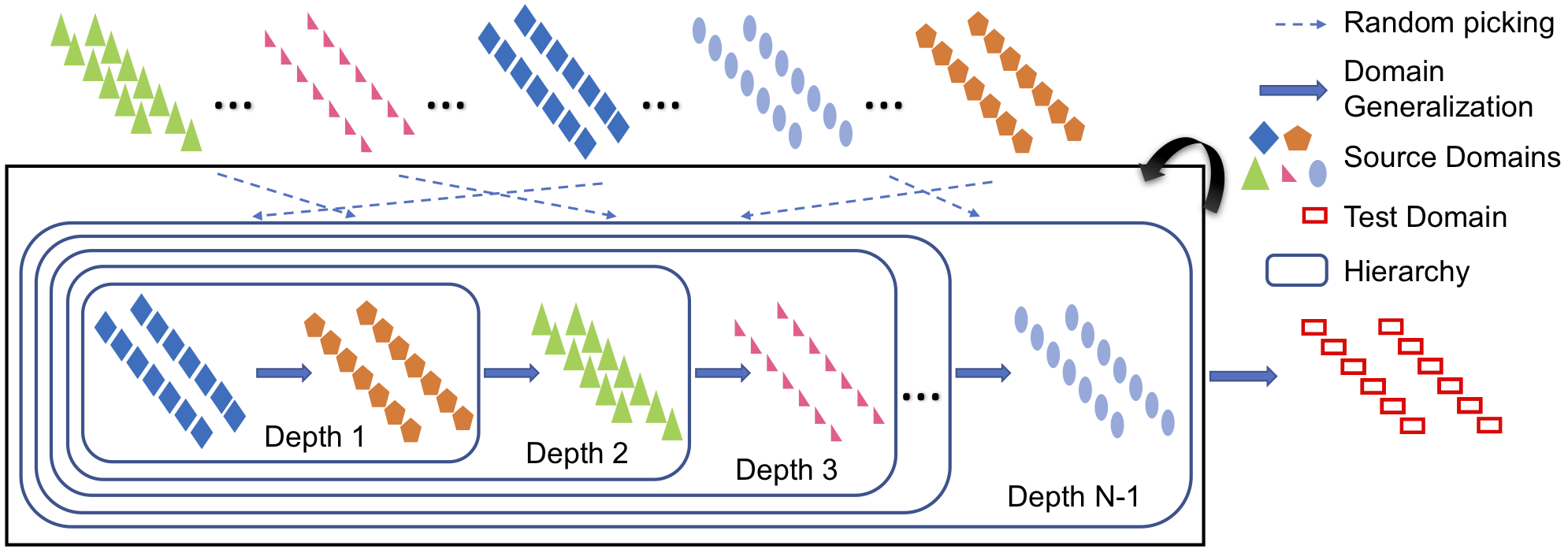}
% \vspace{-0.5cm}
\caption{Schematic illustration of our domain generalization training framework. A base DG method is trained at every step in a sequence of domains. And this is repeated over different random sequences.}
\label{fig:method-illustration}
\end{figure}

\section{Sequential Learning Domain Generalization}
Domain generalization methods mostly aim to achieve $\min\mathcal{L}(\mathcal{D}_*|\mathcal{D}_{\text{train}})$. I.e., low loss on a testing domain $\mathcal{D}_*$ after training on a set of training domains $\mathcal{D}_{\text{train}}$. Of course this can not be optimized in the conventional way since the target $\mathcal{D}_*$ is not available, so various methods \cite{ECCV12_Khosla,ghifary2015domain,muandet2013domaingeneralization} attempt to achieve this indirectly by various kinds of multi-domain training on the domains in $\mathcal{D}_{\text{train}}$. As outlined in the previous section, meta-learning approaches such as MLDG aim to achieve this by finding a model that performs well over many different meta-train and meta-test splits of the true training domains: $\min \mathbb{E}_{(\mathcal{D}_\mathcal{\text{mtrn}},\mathcal{D}_\mathcal{\text{mtst}})\sim \mathcal{D}_{\text{train}}}\mathcal{L}(\mathcal{D}_\mathcal{\text{mtst}}|\mathcal{D}_{\text{mtrn}})$. Inspired by the idea of human lifelong learning-to-learn \cite{smith2002humanL2L} and the benefit of providing `more practice' \cite{doersch2017mtlSelfSup,jaderberg2017unsupAuxRL}, we propose to optimize the performance of a sequentially learned DG model at every step of a trajectory $p$ through the domains, averaged over all possible trajectories $\mathcal{P}$. As illustrated in Fig.~\ref{fig:method-illustration}, this corresponds to:
\small
\begin{equation}
\operatorname{min} \mathbb{E}_{p\sim\mathcal{P}}\sum_{d\in p}\mathcal{L}(\mathcal{D}_{d}|\mathcal{D}_{[:d)})\label{eq:h}
\end{equation}
\normalsize
Here $\mathcal{L}(\mathcal{D}_{d}|\mathcal{D}_{[:d)})$ denotes the performance on meta-test domain $d$ given a DG model which has been \emph{sequentially} trained on meta-train domains before the arrival of domain $d$, and $p$ denotes the sequential trajectory. This covers $N!$ distinct DG learning problems (at each incremental step of each possible trajectory $p$), since the order of the path through any fixed set of source domains matters. The  framework is DG-algorithm agnostic in that does not stipulate which DG algorithm should be used at each step. Any base DG algorithm which can be sequentially updated could be used. In this paper we show how to instantiate this idea for both Undo Bias \cite{ECCV12_Khosla} and MLDG \cite{Li2018MLDG} DG algorithms.

\begin{algorithm}[t]
\SetAlgoLined
\textbf{Input}:$\mathcal{D} = [\mathcal{D}_1, \mathcal{D}_2, \dots, \mathcal{D}_N]$ N source domains.\\
\textbf{Initialize}: $\alpha$, $\beta$ ,$\gamma$ and $\theta$ \\
 \While{not done training}{
%   Randomly sample a trajectory $p$: 
$p=\operatorname{shuffle}([1,2,\dots,N])$ \\
%   Sample a mini-batch $\tilde{\mathcal{D}}_i$ for each domain $\mathcal{D}_i$ \\
  $\tilde{\mathcal{D}} = [\tilde{\mathcal{D}}_1, \tilde{\mathcal{D}}_2, \dots, \tilde{\mathcal{D}}_N]$\hfill//\small{Sample mini-batches $\tilde{\mathcal{D}}_i$}\\
  $\mathcal{L} = \mathcal{L}(\tilde{\mathcal{D}}_{p[1]}, \theta)$ \\
  \For{$i$ \textbf{in} $[2,3,\dots,|p|]$}
  {
  
  $\mathcal{L}$ += $\beta\Big(\mathcal{L}(\tilde{\mathcal{D}}_{p[i]}, \theta - \alpha \nabla_\theta\mathcal{L}) \Big)$\hfill//\small{Inner-loop update}\\
  }
  Update $\theta := \theta -  \gamma\nabla_\theta\mathcal{L} $\hfill//\small{Meta update} \\
  
 }
 \textbf{Output}: $\theta$\ 
 \caption{\nameS: Sequential Learning MLDG}
 \label{alg:full}
\end{algorithm}

% \begin{algorithm}[tb]
% \caption{Hierarchical Domain Generalization}\label{alg:full}
% \begin{algorithmic}[1]
% \State \textbf{Input}: $\mathcal{D} = [\mathcal{D}_1, \mathcal{D}_2, \dots, \mathcal{D}_N]$
% \State \textbf{initialize hyper parameters}: $\lambda$ (for \hundo) or $\alpha$ and $\beta$ (for \nameS{})
% \State \textbf{initialize model parameters}: $[\theta_1,\theta_2,\dots,\theta_N]$ (for \hundo) or $\theta$ (for \nameS{})
% \For{epoch \textbf{in} num\_epochs}
% \State Randomly sample a trajectory $p$ by $p=\operatorname{shuffle}([1,2,\dots,N])$ 
% \For{iteration \textbf{in} num\_iterations}
% \State Randomly sample a mini-batch $\tilde{\mathcal{D}}_i$ for each domain $\mathcal{D}_i$
% \State $\tilde{\mathcal{D}} = [\tilde{\mathcal{D}}_1, \tilde{\mathcal{D}}_2, \dots, \tilde{\mathcal{D}}_N]$
% \State $\mathcal{L} = \mathcal{L}(\tilde{\mathcal{D}}_{p[1]}, \theta_{p[1]})$ (for \hundo) or $\mathcal{L} = \mathcal{L}(\tilde{\mathcal{D}}_{p[1]}, \theta)$ (for \nameS{})
% \For{$i$ \textbf{in} $[2,3,\dots,|p|]$}
% \State $\mathcal{L}$ += $\Big(\mathcal{L}(\tilde{\mathcal{D}}_{p[i]}, \theta_{p[i]}) + \lambda \| \theta_{p[i]} - \frac{\sum_{j=1}^{i-1}\theta_{p[j]}}{i-1} \|_2^2 \Big)$ (for \hundo)
% \State or
% \State $\mathcal{L}$ = $\mathcal{L} + \beta\Big(\mathcal{L}(\tilde{\mathcal{D}}_{p[i]}, \theta - \alpha \nabla_\theta\mathcal{L}) \Big)$ (for \nameS{})
% \EndFor
% \State Do one-step SGD on $\mathcal{L}$
% \EndFor
% \EndFor
% \State \textbf{Output}: $\frac{\sum_{i=1}^{N}\theta_1,\theta_2,\dots,\theta_N}{N}$ (for \hundo) or $\theta$ (for \nameS{})
% \end{algorithmic}
% \end{algorithm}

\subsection{Sequential Learning MLDG (\nameS)}
Vanilla MLDG already optimizes an expectation over meta-train/meta-test splits over the source domains (Section~\ref{sec:mldg}). At every iteration, it randomly samples one domain as meta-test, and keeps the others as meta-train. But within the meta-train domains, it simply aggregates them. It does not exploit their domain grouping. To instantiate our hierarchical training framework (Eq.~\ref{eq:h}) for MLDG we imagine \emph{recursively} applying MLDG. For a given meta-test/meta-train split, we apply MLDG again within the meta-train split until there is only a single domain in the meta-train set. This simulates a lifelong DG learning process, where we should succeed at DG between the first and second training domains, and then the result of that should succeed at DG on the third training domain etc. The objective function to optimize for \nameS{} is:
\small
\begin{equation}
\begin{aligned}
\label{eq:hmldg}
\mathcal{L}_{\text{\nameS{}}} = \mathbb{E}_{p\sim\mathcal{P}}~~~ &\mathcal{L}_1( \mathcal{D}_{p[1]},\theta) \\
+ & \beta \sum_{i=2}^{N} \mathcal{L}_i( \mathcal{D}_{p[i]}, \theta-\alpha\nabla_\theta\sum_{j=1}^{i-1}\mathcal{L}_j)\\
=  \mathbb{E}_{p\sim\mathcal{P}}~~~ &\mathcal{L}_1( \mathcal{D}_{p[1]}, \theta) \\
+ &\beta\mathcal{L}_2(\mathcal{D}_{p[2]}, \theta - \alpha \nabla_\theta\mathcal{L}_1)\\
+ &\beta\mathcal{L}_3(\mathcal{D}_{p[3]}, \theta - \alpha \nabla_\theta\sum_{j=1}^{2}\mathcal{L}_{j}) + ... \\ 
+ &\beta\mathcal{L}_{N}(\mathcal{D}_{{p[N]}}, \theta - \alpha \nabla_\theta\sum_{j=1}^{N-1}\mathcal{L}_{j})
\end{aligned}
\end{equation}
\normalsize
The optimization is carried out over all possible paths $p$ through the training domains.  MLDG is model-agnostic and computes a single parameter $\theta$ for all domains, so the final $\theta$ after optimization is used for inference on unseen domains. The overall algorithm is shown in Alg.~\ref{alg:full}.
% The objective for \nameS{} (Eq.~\ref{eq:hmldg}) is perhaps more intuitive than for \hundo (Eq.~\ref{eq:h-udl}). 
% During training, we sequentially traverse domains, do MLDG training to improve performance on each target domain in the sequence, and then add it to the training set for the next domain. 

\begin{algorithm}[t]
\SetAlgoLined
\textbf{Input}: $\mathcal{D} = [\mathcal{D}_1, \mathcal{D}_2, \dots, \mathcal{D}_N]$ N source domains.\\
\textbf{Initialize}: $\alpha$, $\beta$, $\gamma$ and $\theta$\\
\While{not done training}{
 $\tilde{\theta} = \theta$\\
 $p=\operatorname{shuffle}([1,2,\dots,N])$ \\
%   Sample a mini-batch $\tilde{\mathcal{D}}_i$ for each domain $\mathcal{D}_i$ \\
  $\tilde{\mathcal{D}} = [\tilde{\mathcal{D}}_1, \tilde{\mathcal{D}}_2, \dots, \tilde{\mathcal{D}}_N]$\hfill//\small{Sample mini-batches $\tilde{\mathcal{D}}_i$}\\
\For{$i$ \textbf{in} $[1, N]$}
{
$\mathcal{L}_i = \beta\mathcal{L}(\tilde{\mathcal{D}}_{p[i]}, \tilde{\theta}) $\\ 
 $\tilde{\theta} = \tilde{\theta} - \alpha \nabla_{\tilde{\theta}} \mathcal{L}_i$\hfill//\small{Inner-loop update} \\ 
}
 Update $\theta :=  \theta + \gamma(\tilde{\theta} - \theta)$\hfill//\small{Meta update}\\
}
\textbf{Output}: $\theta$
 \caption{Faster First-Order \nameS{}}
 \label{alg:reptile-dg}
\end{algorithm}
%\keypoint{Analysis of \nameS{}} 
% \subsubsec{Analysis of \nameS{}:} 
\keypoint{MLDG}
The MLDG mechanism was originally analyzed \cite{Li2018MLDG} via a first-order Taylor series. Since MLDG only does one-step DG validation, one domain is sampled as meta-test to split the source domains. Then the objective function is

\small
\begin{equation}
\begin{aligned}
\label{eq:mldg}
\mathcal{L}_{\text{MLDG}} = & \mathcal{L}_1(\mathcal{D}_{\text{mtrn}},\theta) +  \beta\mathcal{L}_2( \mathcal{D}_{\text{mtst}}, \theta-\alpha\nabla_\theta\mathcal{L}_1)
\end{aligned}
\end{equation}
\normalsize
After Taylor expansion on the second item, it  becomes
\small
\begin{equation}
\begin{aligned}
\label{eq:mldg-sec-item-taylor}
\mathcal{L}_2(\theta-\alpha\nabla_\theta\mathcal{L}_1) = \mathcal{L}_2(\theta) + \nabla_\theta\mathcal{L}_2 \cdot(-\alpha\nabla_\theta\mathcal{L}_1)
\end{aligned}
\end{equation}
\normalsize
and then $\mathcal{L}_{\text{MLDG}}$ becomes
\small
\begin{equation}
\begin{aligned}
\label{eq:mldg-taylor}
\mathcal{L}_{\text{MLDG}} = \mathcal{L}_1(\theta) +  \beta \mathcal{L}_2(\theta) - \beta \alpha \nabla_\theta\mathcal{L}_1 \nabla_\theta\mathcal{L}_2
\end{aligned}
\end{equation}
\normalsize
\noindent This led to MLDG's interpretation as a preference for an optimization path with \emph{aligned} gradients between meta-train and meta-test \cite{Li2018MLDG}.

\keypoint{\nameS{}}
If we use 3 source domains as an example to analyse \nameS{}, the loss function is
\small
\begin{equation}
\begin{aligned}
\label{eq:hmldg-3s}
 \mathcal{L}_{\text{\nameS{}-3}}   = & \mathcal{L}_1(\theta) + \beta\mathcal{L}_2(\theta-\alpha\nabla_\theta\mathcal{L}_1) \\
 & + \beta\mathcal{L}_3(\theta-\alpha\nabla_\theta(\mathcal{L}_1 +\mathcal{L}_2))
%  &  = \mathcal{L}_1(\theta) +  \beta \mathcal{L}_2(\theta) - \beta \alpha \nabla_\theta\mathcal{L}_1 \nabla_\theta\mathcal{L}_2 \\
% &  ~~~~ + \beta \mathcal{L}_3(\theta) - \beta\alpha \nabla_\theta\mathcal{L}_3 ( \nabla_\theta\mathcal{L}_1 + \nabla_\theta\mathcal{L}_2) \\
% &  = \mathcal{L}_1(\theta) +  \beta \mathcal{L}_2(\theta) + \beta \mathcal{L}_3(\theta) \\
% &  ~~~~  - \beta \alpha \nabla_\theta\mathcal{L}_1 \nabla_\theta\mathcal{L}_2 - \beta\alpha \nabla_\theta\mathcal{L}_3 \nabla_\theta\mathcal{L}_1 - \beta\alpha \nabla_\theta\mathcal{L}_3\nabla_\theta\mathcal{L}_2
\end{aligned}
\end{equation}
\normalsize
The first two items are the same as $\mathcal{L}_{\text{MLDG}}$. Apply Taylor expansion on the third item in $\mathcal{L}_{\text{\nameS{}-3}}$,
\begin{equation}
    \mathcal{L}_3(\theta-\alpha\nabla_\theta(\mathcal{L}_1 +\mathcal{L}_2)) = \mathcal{L}_3(\theta) + \nabla_\theta\mathcal{L}_3 \cdot(-\alpha\nabla_\theta(\mathcal{L}_1 +\mathcal{L}_2))
\end{equation}
we have,
\small
\begin{equation}
\begin{aligned}
\label{eq:hmldg-3s}
 \mathcal{L}_{\text{\nameS{}-3}} &  =  \mathcal{L}_1(\theta) +  \beta \mathcal{L}_2(\theta) + \beta \mathcal{L}_3(\theta) \\
& \hspace{-1cm} - \beta \alpha \nabla_\theta\mathcal{L}_1 \nabla_\theta\mathcal{L}_2 - \beta\alpha \nabla_\theta\mathcal{L}_3 \nabla_\theta\mathcal{L}_1 - \beta\alpha \nabla_\theta\mathcal{L}_3\nabla_\theta\mathcal{L}_2
\end{aligned}
\end{equation}
\normalsize
This shows that \nameS{} optimizes all source domains (first three terms), while preferring an optimization path where gradients align across all pairs of domains (second three terms maximising dot products). This is different to MLDG, that only optimizes the inner product of gradients between the current  meta-train and meta-test domain splits. In contrast \nameS{} has the chance to optimize for DG on each meta-test domain in the sequential way, thus obtaining more unique experience to `practice' DG.

%of `practicing' DG on each unseen meta-test domain in lifelong learning sequences. And in each DG `practice' it does MLDG once, which results in learning the inner product between all traversed domains.

\keypoint{A Direct \nameS{} Implementation}
A direct implementation of the meta update for \nameS{} in the three domain case would differentiate $ \mathcal{L}_{\text{\nameS{}-3}}$ (Eq.~\ref{eq:hmldg-3s}) w.r.t $\theta$ as
\small
\begin{equation}
\begin{aligned}
\label{eq:hmldg-3s-grad}
\nabla_\theta\mathcal{L}_{\text{\nameS{}-3}} & = \nabla_\theta\mathcal{L}_1(\theta) + \beta\nabla_\theta\mathcal{L}_2(\theta-\alpha\nabla_\theta\mathcal{L}_1) \\
& ~~~~~ + \beta\nabla_\theta\mathcal{L}_3(\theta-\alpha\nabla_\theta(\mathcal{L}_1 +\mathcal{L}_2)) \\
 & = \frac{\partial \mathcal{L}_1(\theta)}{\partial \theta} + \beta \frac{\partial \mathcal{L}_2(\theta_1)}{\partial \theta_1}\frac{\partial \theta_1}{\partial \theta} + \beta\frac{\partial \mathcal{L}_3(\theta_2)}{\partial \theta_2}\frac{\partial \theta_2}{\partial \theta} \\
\end{aligned}
\end{equation}
\normalsize
where
\small
\begin{equation}
\label{eq:theta1}
\begin{aligned}
\theta_1&= \theta - \alpha\nabla_\theta\mathcal{L}_1\\
\cut{\end{aligned}
\end{equation}
\normalsize
and
\small
\begin{equation}
\label{eq:theta2}
\begin{aligned}}
\theta_2 &= \theta-\alpha\nabla_\theta(\mathcal{L}_1 +\mathcal{L}_2) \\
\end{aligned}
\end{equation}
\normalsize

% \begin{algorithm}[t]
% \caption{Faster First-order \nameS{}}\label{alg:reptile-dg}
% \begin{algorithmic}[1]
% \State \textbf{Input}: $\mathcal{D} = [\mathcal{D}_1, \mathcal{D}_2, \dots, \mathcal{D}_N]$ n souce domains.
% \State Initialize the initial parameters $\theta$, inner step size $\alpha$ and outer step size $\gamma$
% \For{iteration = 1, 2, ...}
% \State $\tilde{\theta} = \theta$
% \State $p = \textnormal{Shuffle} ([1, 2, ...,N])$
% \For{$i$ in [1, N]}
% \State $\tilde{\theta} = \tilde{\theta} - \alpha \nabla_{\tilde{\theta}} \mathcal{L}(\mathcal{D}_{p[i]}, \tilde{\theta})$
% \EndFor
% \State Update $\theta \leftarrow \theta + \gamma(\tilde{\theta} - \theta)$
% \EndFor
% \end{algorithmic}
% \end{algorithm}

\noindent However, update steps based on Eq.~\ref{eq:hmldg-3s-grad} require high-order gradients when computing $\frac{\partial \theta_1}{\partial \theta}$, $\frac{\partial \theta_2}{\partial \theta} $. These higher-order gradients are expensive to compute. 

\subsection{First-order Approximator of \nameS{}} 
%\keypoint{A Naive First Order Approximation (\nameFS{})} 
\keypoint{\nameFS{}:} 
Similar to \cite{finn2017model}, we can alleviate the above issue by stopping the gradient of the exposed first derivative items to omit higher-order gradients. I.e, $\nabla_\theta\mathcal{L}_1$ and $\nabla_\theta(\mathcal{L}_1 +\mathcal{L}_2)$ in Eq.~\ref{eq:theta1} are constants when computing $\mathcal{L}_2$ and $\mathcal{L}_3$. Then for \nameFS{}, Eq.~\ref{eq:hmldg-3s-grad} becomes
% To save computational efforts, we use this approximator for \nameS{} by default.
\small
\begin{equation}
\begin{aligned}
\label{eq:hmldg-3s-fo}
\nabla_\theta\mathcal{L}_{\text{\nameS{}-3}} & = \frac{\partial \mathcal{L}_1(\theta)}{\partial \theta} + \beta \frac{\partial \mathcal{L}_2(\theta_1)}{\partial \theta_1}\frac{\partial \theta_1}{\partial \theta} + \beta\frac{\partial \mathcal{L}_3(\theta_2)}{\partial \theta_2}\frac{\partial \theta_2}{\partial \theta} \\
& = \frac{\partial \mathcal{L}_1(\theta)}{\partial \theta} + \beta \frac{\partial \mathcal{L}_2(\theta_1)}{\partial \theta_1}\frac{\partial (\theta -\alpha \nabla_{\theta}\mathcal{L}_1 )}{\partial \theta} \\
&~~~~~+ \beta\frac{\partial \mathcal{L}_3(\theta_2)}{\partial \theta_2}\frac{\partial (\theta-\alpha\nabla_\theta(\mathcal{L}_1 +\mathcal{L}_2))}{\partial \theta}\\
& = \frac{\partial \mathcal{L}_1(\theta)}{\partial \theta} + \beta \frac{\partial \mathcal{L}_2(\theta_1)}{\partial \theta_1} + \beta\frac{\partial \mathcal{L}_3(\theta_2)}{\partial \theta_2}
% & = \mathcal{L}_1^{'} + \beta \mathcal{L}_2^{'} + \beta \mathcal{L}_3^{'} \\
\end{aligned}
\end{equation}
\normalsize
\nameFS{} still follows Alg.~\ref{alg:full}, but saves computation by omitting high-order gradients in back propagation. We use this approximator for \nameS{} by default.
However \nameFS{} still requires back propagation (as per Eq.~\ref{eq:hmldg-3s-grad}), to compute gradients of $\mathcal{L}_1$, $\mathcal{L}_2$ and $\mathcal{L}_3$, even though higher-order gradients are ignored.
%respectively.}

\subsection{Fast First-Order \nameS{}}
%As explained above \nameFS{} saves computation by omitting high-order gradients in back propagation. \doublecheck{However \nameFS{} still requires back propagation to compute gradients of $\mathcal{L}_1$, $\mathcal{L}_2$ and $\mathcal{L}_3$ respectively.}
\keypoint{\nameFFS{}:} If we look at $\nabla_\theta\mathcal{L}_{\text{\nameS{}-3}}$ in Eq.~\ref{eq:hmldg-3s-fo} again, we find
\small
\begin{equation}
\begin{aligned}
\label{eq:hmldg-3s-ffo}
\frac{\partial \mathcal{L}_1(\theta)}{\partial \theta} + \beta \frac{\partial \mathcal{L}_2(\theta_1)}{\partial \theta_1} + \beta\frac{\partial \mathcal{L}_3(\theta_2)}{\partial \theta_2} & = \mathcal{L}_1^{'} + \beta \mathcal{L}_2^{'} + \beta \mathcal{L}_3^{'}
% & = \mathcal{L}_1^{'} + \beta \mathcal{L}_2^{'} + \beta \mathcal{L}_3^{'} \\
\end{aligned}
\end{equation}
\normalsize
This means that one-step meta update of \nameFS{} is $\gamma (\mathcal{L}_1^{'} + \beta \mathcal{L}_2^{'} + \beta \mathcal{L}_3^{'})$, where $\gamma$ is the meta step-size. This indicates that one update of naive first-order \nameS{} is equivalent to updating the parameters towards the result of training on $\mathcal{L}_1(\mathcal{D}_{p[1]})$, $\mathcal{L}_2(\mathcal{D}_{p[2]})$ and $\mathcal{L}_3(\mathcal{D}_{p[3]})$ recursively. In other words, if we regard the initial parameters as $\theta$ and the parameters updated recursively on $\mathcal{L}_1(\mathcal{D}_{p[1]})$, $\mathcal{L}_2(\mathcal{D}_{p[2]})$ and $\mathcal{L}_3(\mathcal{D}_{p[3]})$ as $\tilde{\theta}$, then Eq.~\ref{eq:hmldg-3s-ffo} can be expressed as
\small
\begin{equation}
\begin{aligned}
\label{eq:hmldg-3s-ffo-interpolation}
\mathcal{L}_1^{'} + \beta \mathcal{L}_2^{'} + \beta \mathcal{L}_3^{'} & = \theta - \tilde{\theta}  \\
% & = \mathcal{L}_1^{'} + \beta \mathcal{L}_2^{'} + \beta \mathcal{L}_3^{'} \\
\end{aligned}
\end{equation}
\normalsize

This means that we can optimize $\mathcal{L}_1(\mathcal{D}_{p[1]})$, $\mathcal{L}_2(\mathcal{D}_{p[2]})$ and $\mathcal{L}_3(\mathcal{D}_{p[3]})$ in sequence (to obtain $\tilde{\theta}$, and then use the resulting offset vector as the meta-gradient for updating $\theta$). Thus we do not need to backpropagate to explicitly compute the gradients suggested by Eq.~\ref{eq:hmldg-3s-ffo}. 
%i.e. we can use the interpolation vector between $\tilde{\theta}$ and $\theta$ as the gradient to update $\theta$. In this way, we can use the right-hand side in Eq.~\ref{eq:hmldg-3s-ffo-interpolation} as our meta gradient and we do not need to back propagate our model to explicitly compute gradients by the left-hand side. This saves computation of model update especially when we use deep neural networks, whose back propagation takes time. 
The overall flow of \nameFFS{} is shown in Alg.~\ref{alg:reptile-dg}.

% We also did the theoretical analysis of this faster first-order \nameS{} (FFO-\nameS{}) shown in Appendix \ref{appendix-FFO-HMLDG}. The analysis proves that FFO-\nameS{} learns to maximize the inner-product between gradients of different domains in expectation of training iterations, which is slightly different to (first-order or) \nameS{} doing that in per iteration.

\keypoint{Link between Fast First-Order \nameS{} and \nameS{}}
We analyze \nameFFS{} in Alg.~\ref{alg:reptile-dg} considering two source domains and derive the expectation of the optimization gradient is 
\small
\begin{equation}
\label{eq:ffo-hmldg-expectation-two-loss}
\begin{aligned}
 \mathbb{E}_{p\sim\mathcal{P}}[g_{p[1]}+g_{p[2]}] = \bar{g}_1 + \bar{g}_2 - \frac{\alpha}{2} \frac{\partial (\bar{g}_1 \cdot \bar{g}_2)}{\partial \tilde{\theta}_1} + \mathcal{O}(\alpha^2)\\
\end{aligned}
\end{equation}
\normalsize
% \small
% \begin{equation}
%     \begin{aligned}
%     \frac{1}{2} (- \alpha \bar{H}_1 \bar{g}_2 - \alpha \bar{H}_2 \bar{g}_1 ) = - \frac{\alpha}{2} \frac{\partial (\bar{g}_1 \cdot \bar{g}_2)}{\partial \tilde{\theta}_1}
%     \end{aligned}
% \end{equation}
% \normalsize
Here $\bar{g}_1$, $\bar{g}_2$ are the gradient updates for the first and second source domains and $\bar{g}_1 \cdot \bar{g}_2$ is the inner product between the two gradients. The gradient $ - \frac{\partial (\bar{g}_1 \cdot \bar{g}_2)}{\partial \tilde{\theta}_1}$ is in the direction that maximizes it. This means in expectation of multiple gradient updates \nameFFS{} learns to maximize the inner-product between gradients of different domains. Thus it maintains a similar but slightly different objective to \nameS{}, which maximizes the inner-product of gradients in each meta update. More details can be found in Appendix~\ref{app-analysis-ffo}.

\section{Experiments}

% \begin{table*}[tb]
% \centering
% \scalebox{0.522}{
% \begin{tabular}{cc|cccccccccccccc}
% \toprule
% Source  & Target  & DICA \cite{muandet2013domaingeneralization} & LRE-SVM \cite{Xu2014lre} &D-MTAE \cite{ghifary2015domain} & CCSA \cite{motiian2017CCSA}&MMD-AAE \cite{mmdaaecvpr2018} & DANN\cite{ganin2016dann} &AGG & CrossGrad \cite{shiv2018dg} & MetaReg \cite{NIPS2018_metareg}  & Undo-Bias~\cite{ECCV12_Khosla} & \hundo & MLDG \cite{Li2018MLDG}& FFO-\nameS{} & \nameS{}  \\ \midrule
% % \hline
% 0,1,2,3&4& 61.5 & 75.8& 78.0& 75.8& 79.1&  81.6  & 80.0 & 78.4 &79.3&   80.7  &82.7&79.4&   81.1       &80.1         \\
% 0,1,2,4&3& 72.5& 86.9& 92.3& 92.3& 94.5 &  94.5  & 94.5& 94.2 &94.5&   95.3  &94.9&95.2&   95.1       &95.0         \\
% 0,1,3,4&2& 74.7& 84.5& 91.2& 94.5& 95.6 &   100.0~~~  &99.8& 100.0~~~&99.8&   99.9   &100.0~~~&99.9& 100.0~~~& 99.8                   \\
% 0,2,3,4&1& 67.0& 83.4& 90.1& 91.2& 93.4 &  91.7  & 93.4& 94.0 &92.5&   94.8  &94.0&95.2&  93.1     &96.2               \\
% 1,2,3,4&0& 71.4& 92.3& 93.4& 96.7& 96.7 &  93.5 & 93.5 & 91.2 &92.8&    94.2 &93.9&90.4&  94.3    & 92.7                    \\
% \midrule
% \multicolumn{2}{c|}{Ave.} & 69.4& 84.6& 87.0& 90.1& 91.9& 92.3 & 92.2 & 91.6 &91.8&  93.0  &93.1&91.9&92.7 &92.8    \\
% \bottomrule
% \end{tabular}
% }
% \vspace{-0.3cm}
% \caption{\small Cross-view action recognition results (accuracy. \%) on IXMAS dataset.}
% \label{tab:ixmax}
% \end{table*}

\begin{table*}[tb]
\centering
\caption{\small Performance on IXMAS action recognition. Leave one camera-view out. Accuracy (\%).}
% \vspace{-0.2cm}
\resizebox{1.0\textwidth}{!}{
\begin{tabular}{c|cccccccccc}
\toprule
 Unseen  & \textbf{MMD-AAE} \cite{mmdaaecvpr2018} &\textbf{AGG} & \textbf{DANN}\cite{ganin2016dann} & \textbf{CrossGrad} \cite{shiv2018dg} & \textbf{MetaReg} \cite{NIPS2018_metareg}  & \textbf{Undo-Bias}~\cite{ECCV12_Khosla} & \textbf{\hundo} & \textbf{MLDG} \cite{Li2018MLDG}& \textbf{\nameFFS{}} & \textbf{\nameS{}}  \\ \toprule
% \hline
4&79.1 &  80.0   & 81.6 & 78.4 &79.3&   80.7  &82.7&79.4&   81.1       &80.1         \\
3& 94.5 & 94.5  &94.5 & 94.2 &94.5&   95.3  &94.9&95.2&   95.1       &95.0         \\
2&95.6   & 99.8&100.0~~~& 100.0~~~&99.8&   99.9   &100.0~~~&99.9& 100.0~~~& 99.8                   \\
1&93.4  & 93.4  &91.7 & 94.0 &92.5&   94.8  &94.0&95.2&  93.1     &96.2               \\
0& 96.7 & 93.5 & 93.5 & 91.2 &92.8&    94.2 &93.9&90.4&  94.3    & 92.7                    \\
\bottomrule
Ave. &91.9 & 92.2 & 92.3 & 91.6 &91.8&  93.0  &93.1&91.9&92.7 &92.8    \\
\bottomrule
\end{tabular}
}

\label{tab:ixmax}
\end{table*}

% \begin{table*}[t]
% \centering
% \scalebox{0.53}{
% \begin{tabular}{c|ccccccccccccccc}
% % \toprule
%  Target & \textbf{DICA} \cite{muandet2013domaingeneralization}& \textbf{LRE-SVM} \cite{Xu2014lre} & \textbf{D-MTAE} \cite{ghifary2015domain}& \textbf{CCSA} \cite{motiian2017CCSA}& \textbf{MMD-AAE}\cite{mmdaaecvpr2018}& \textbf{DANN} \cite{ganin2016dann}  & \textbf{AGG} & \textbf{CrossGrad}~\cite{shiv2018dg}& \textbf{MetaReg}~\cite{NIPS2018_metareg} & \textbf{Undo-Bias}\cite{ECCV12_Khosla} & \textbf{HUndo-Bias} & \textbf{MLDG} \cite{Li2018MLDG} & \textbf{FFO-\nameS{}} & \textbf{HMLDG}\\
% \toprule
% V& 63.7& 60.6& 63.9& 67.1& 67.7& 66.4    & 65.4  &65.5 &65.0 &68.1 & 68.7 &67.7&68.1& 68.7\\
% L& 58.2& 59.7& 60.1& 62.1& 62.6& 64.0    & 60.6  &60.0 &60.2 &60.3 & 61.8&61.3&63.1& 64.8\\
% C& 79.7& 88.1& 89.1& 92.3& 94.4& 92.6    & 93.1  &92.0 &92.3 &93.7 & 95.0&94.4&94.8& 96.4\\
% S& 61.0& 54.9& 61.3& 59.1& 64.4& 63.6    & 65.8  &64.7 &64.2&66.0 & 66.1&65.9&65.2& 64.0\\
% \toprule
% \multicolumn{1}{c|}{Ave.}& 65.7& 65.8& 68.6& 70.2& 72.3& 71.7  & 71.2  &70.5 &70.4 & 72.0 & 72.9& 72.3&72.8& 73.5\\
% % \bottomrule
% \end{tabular}
% }
% \vspace{-0.3cm}
% \caption{\small Cross-dataset object recognition results (accuracy. \%) on VLCS benchmark.}
% \label{tab:vlcs}
% \end{table*}

\begin{table*}[t]
\centering
\caption{Performance on VLCS object recognition. Leave one dataset out. Accuracy (\%).}
% \vspace{-0.2cm}
\resizebox{1.0\textwidth}{!}{
\begin{tabular}{c|ccccccccccc}
\toprule
 Unseen & \textbf{MMD-AAE}\cite{mmdaaecvpr2018}  & \textbf{AGG}& \textbf{DANN} \cite{ganin2016dann} & \textbf{CrossGrad}~\cite{shiv2018dg}& \textbf{MetaReg}~\cite{NIPS2018_metareg} & \textbf{Undo-Bias}\cite{ECCV12_Khosla} & \textbf{\hundo} & \textbf{MLDG} \cite{Li2018MLDG} & \textbf{\nameFFS{}} & \textbf{\nameS{}}\\
\toprule
V&  67.7&  65.4  &  66.4  &65.5 &65.0 &68.1 & 68.7 &67.7&68.1& 68.7\\
L&  62.6&  60.6  &  64.0  &60.0 &60.2 &60.3 & 61.8&61.3&63.1& 64.8\\
C&  94.4&  93.1  &  92.6  &92.0 &92.3 &93.7 & 95.0&94.4&94.8& 96.4\\
S&  64.4&  65.8  &  63.6  &64.7 &64.2&66.0 & 66.1&65.9&65.2& 64.0\\
\bottomrule
\multicolumn{1}{c|}{Ave.}& 72.3&  71.2 & 71.7  &70.5 &70.4 & 72.0 & 72.9& 72.3&72.8& 73.5\\
\bottomrule
\end{tabular}
}
\label{tab:vlcs}
\end{table*}

\begin{table*}[t]
\centering
\caption{\small Performance on PACS object recognition across styles (ResNet-18). Accuracy (\%).}
% \vspace{-0.2cm}
\resizebox{1.\textwidth}{!}{
\begin{tabular}{c|ccccccccc}
\toprule
Unseen & \textbf{AGG} & \textbf{DANN} \cite{ganin2016dann} & \textbf{CrossGrad} \cite{shiv2018dg}& \textbf{MetaReg}~\cite{NIPS2018_metareg} & \textbf{Undo-Bias} \cite{ECCV12_Khosla}& \textbf{\hundo}  & \textbf{MLDG} \cite{Li2018MLDG} & \textbf{\nameFFS{}} & \textbf{\nameS{}} \\ \toprule
A & 77.6 & 81.3 & 78.7  &79.5 & 78.4 & 80.6 & 79.5 & 80.0 & 80.5 \\
C & 73.9 & 73.8 & 73.3  &75.4 & 72.5 & 76.2 & 77.3 & 77.4 & 77.8 \\
P & 94.4 & 94.0 & 94.0  &94.3 & 92.8 & 94.1 & 94.3 & 94.6 & 94.8 \\
S & 70.3 & 74.3 & 65.1  &72.2 & 73.3 & 72.2 & 71.5 & 73.8 & 72.8 \\
\toprule
Ave. & 79.1 & 80.8 & 77.8  &80.4 & 79.3 & 80.8 & 80.7 & 81.4 & 81.5 \\
\bottomrule
\end{tabular}
}
% %\vspace{-0.6cm}
    \label{tab:agg-resnet-pacs}
\end{table*}

\keypoint{Datasets and Settings}
We evaluate our method on three different benchmarks: \textbf{IXMAS}~\cite{daniel2006ixmax}, where human actions are recognized across different camera views. \textbf{VLCS}~\cite{chen2013vlcs}, which requires the domain generalization across different photo datasets. And \textbf{PACS}~\cite{da2017dg} which is a more realistic and challenging cross-domain visual benchmark of images with different style depictions.
% Finally, \textbf{Visual Decathalon}~\cite{rebuff2017mdlvisdecath} is a major benchmark that aggregates 10 existing visual recognition datasets. Decathalon was originally proposed for multi-domain deep learning (the challenge of learning a single model that simultaneously performs well on all domains in the decathlon). In this paper we go one step beyond this and repurpose Decathalon for the even more challenging task of deep DG. This means that a method should not only perform well on multiple training domains, but generalize to domains within the decathalon that are held out from training. 

\keypoint{Competitors}
For comparative evaluation we also evaluate the following competitors:
\begin{itemize}[noitemsep]%,nolistsep]
    \item \textbf{AGG:} A simple but effective baseline of aggregating all source domains' data for training \cite{da2017dg}. 
    \item \textbf{DANN:} Domain adversarial neural networks learns a domain invariant representation such that source domains cannot be distinguished \cite{ganin2016dann}. 
    \cut{\item \textbf{TF-CNN:} \cite{da2017dg} learns a domain-agnostic model by factoring out the common component from a set of domain-specific models, as well as tensor factorization to compress the model parameters.
    \item \textbf{Fusion:} \cite{Massimiliano2018ICIP} learns the domain-specific source domains and fuses them for testing on the target domain. }
    \item \textbf{MMD-AAE:} A recent DG method which combines kernel MMD and the adversarial auto encoder \cite{mmdaaecvpr2018}. 
    \item \textbf{CrossGrad:} A recently proposed strategy that learns the manifold of training domains, and uses cross-gradients to generate synthetic data that helps the classifier generalize across the manifold \cite{shiv2018dg}.
    \item \textbf{MetaReg:} A latest DG method by meta-learning a regularizer constraining the model parameters to be more domain-generalizable \cite{NIPS2018_metareg}.
    \item \textbf{Undo-Bias:} Undo-Bias models \cite{ECCV12_Khosla} each training domain as a linear combination of a domain-agnostic model and domain-specific bias, and then uses the domain-agnostic model for testing. We use the vanilla deep generalization of Undo-Bias explained in \cite{da2017dg}.
    \item \textbf{MLDG:} A recent DG method that is model-agnostic and meta-learns the domain-generalizable model parameters.
\end{itemize}
The most related alternatives are \textbf{Undo Bias} \cite{ECCV12_Khosla} and \textbf{MLDG} \cite{Li2018MLDG}, which are the models we extend to realize our sequential learning strategy. We re-implement AGG, DANN, CrossGrad, MetaReg, Undo-Bias and MLDG; and report the numbers stated by \cut{TF-CNN, Fusion and} MMD-AAE.

\subsection{Action Recognition Across Camera Views}
\keypoint{Setup}
The IXMAS dataset contains 11 different human actions recorded by 5 video cameras with different views (referred as 0,...,4). The goal is to train an action recognition model on a set of source views (domains), and recognize the action from a novel target view (domain).
We follow \cite{mmdaaecvpr2018} to keep the first 5 actions and use the same Dense trajectory features as input. For our implementation, we follow \cite{mmdaaecvpr2018} to use a one-hidden layer MLP with 2000 hidden neurons as backbone and report the average of 20 runs. In addition, we normalize the hidden embedding by BatchNorm \cite{pmlr-v37-ioffe15} as this gives a good start point for AGG.

\keypoint{Results}
From the results in Table~\ref{tab:ixmax}, we can see that several recent DG methods \cite{Li2018MLDG,shiv2018dg,NIPS2018_metareg,mmdaaecvpr2018} fail to improve over the strong AGG baseline. Undo-Bias works here, and provides $0.8\%$ improvement. Our extension \hundo{} provides a modest increase of $0.1\%$ on the overall accuracy over vanilla Undo. While the original MLDG \cite{Li2018MLDG} fails to improve on the AGG baseline, our \nameS{} provides a 0.9\% gain over MLDG and thus improves 0.6\% on AGG. Our \nameFFS{} runs on par with \nameS{}, demonstrating the efficacy of our approximator. Overall our \hundo{}, \nameFFS{} and \nameS{} all provide a gain in performance over the AGG baseline.

%Sadly, \cite{Li2018MLDG} per se does not work well in this scenario, which is slightly worse than the simple AGG. But interestingly, \nameS{} makes $0.9\%$ performance gain over MLDG, which results $0.6\%$ accuracy better than AGG. Delightfully the faster first-order \nameS{} runs with the performance on par with \nameS{}, which demonstrates the effectiveness of this approximator. Overall, our \hundo, FFO-\nameS{} and \nameS{} all provide the performance gains to the base DG methods and demonstrate the state-of-the-art performances.

\subsection{Object Recognition Across Photo Datasets}
\keypoint{Setup} VLCS domains share 5 categories: bird, car, chair, dog and person. We use  pre-extracted DeCAF6 features and follow \cite{motiian2017CCSA} to randomly split each domain into train (70\%) and test (30\%) and do leave-one-out evaluation. We use a 2 fully connected layer architecture with output size of 1024 and 128 with ReLU activation, as per \cite{motiian2017CCSA} and report the average performance of 20 trials. 
% The optimizer is M-SGD with learning rate 1e-3, momentum 0.9 and weight decay 5e-5. 
% \todo{What is the architecture? End to-end? fixed MLP like IXMAX?}

\keypoint{Results}
In this benchmark, the results in Table~\ref{tab:vlcs} show that the simple AGG method works well again. Recent DG methods \cite{shiv2018dg,NIPS2018_metareg} still struggle to beat this baseline. The base DG methods Undo-Bias \cite{ECCV12_Khosla} and MLDG \cite{Li2018MLDG} work well here, producing comparable results to the state-of-the-art \cite{mmdaaecvpr2018}. Our extensions of these base DG methods, \hundo{}, \nameFFS{} and \nameS{} all provide improvements. Overall our \nameS{} performs best, followed closely by \hundo{} and \nameFFS{}.

% \begin{minipage}{\linewidth}
% \begin{minipage}[b]{0.8\linewidth}
% \begin{figure}[H]
%     \centering
%     \includegraphics[width=0.24\linewidth]{figure/feat1.png}
%     \includegraphics[width=0.24\linewidth]{figure/feat2.png}
%     \includegraphics[width=0.24\linewidth]{figure/fc.png}
%     \caption{Weight similarities of different branches in Undo-Bias and HUndo-Bias. Left to right: first to third layer.}
%     \label{fig:hundo-vs-undo}
% \end{figure}
% \end{minipage}
% \begin{minipage}[b]{0.1\linewidth}
% \end{minipage}
% \end{minipage}

% \begin{table}[!h]
%     \centering
%     \scalebox{0.8}{
%     \begin{tabular}{l|cccc|c}
%     \toprule
%          & V &L & C& S& Ave.  \\
%          \midrule
%         Naive ensemble & 57.3 & 60.4 & 71.5 & 52.6 & 60.5  \\
%         HUndo-Bias & 68.7 & 61.8& 95.0& 66.1& 72.9\\
%          \bottomrule
%     \end{tabular}
%     }
%     \caption{HUndo-Bias \textit{vs} Naive ensemble}
%     \label{tab:hundo-vs-fusion}
% \end{table}

\subsection{Object Recognition Across Styles}
\keypoint{Setup}
The PACS benchmark \cite{da2017dg} contains 4 domains: photo, art painting, cartoon and sketch and 7 common categories: `dog', `elephant', `giraffe', `guitar', `horse', `house' and `person'. \cite{da2017dg} showed that this benchmark has much stronger domain shift than others such as Caltech-Office and VLCS. We use a ResNet-18 pre-trained ImageNet as a modern backbone for comparison. We note that MetaReg~\cite{NIPS2018_metareg} used a slightly different setup than the official PACS protocol \cite{da2017dg}, for which their AGG baseline is hard to reproduce. So we stick to the official protocol and rerun MetaReg. To save computational cost, since Undo-Bias and \hundo{} require domain-specific branches that are expensive when applied to ResNet, we only apply these methods to the last ResNet-18 layer -- so previous layers are shared as per AGG.

%In this setting, we follow \cite{da2017dg} in using the ImageNet pretrained AlexNet, and also compare a ResNet-18 pretrained on ImageNet a more modern backbone for further comparison. To save computational cost, since Undo-Bias and \hundo{} require domain-specific branches, we only apply them to the last layer -- so previous layers are shared as per AGG.

%shown in Table~\ref{tab:agg-alex} and also use a modern ResNet-18 pretrained on ImageNet as another backbone for further comparisons shown in Table~\ref{tab:agg-resnet-pacs}. And Undo-Bias requires domain specific branches for training. To save computational cost, in this deep networks we only apply Undo-Bias and \hundo to the last layer, i.e. the previous layers are shared between domains.

\begin{figure}[t]
    \centering
    \includegraphics[width=1.0\columnwidth]{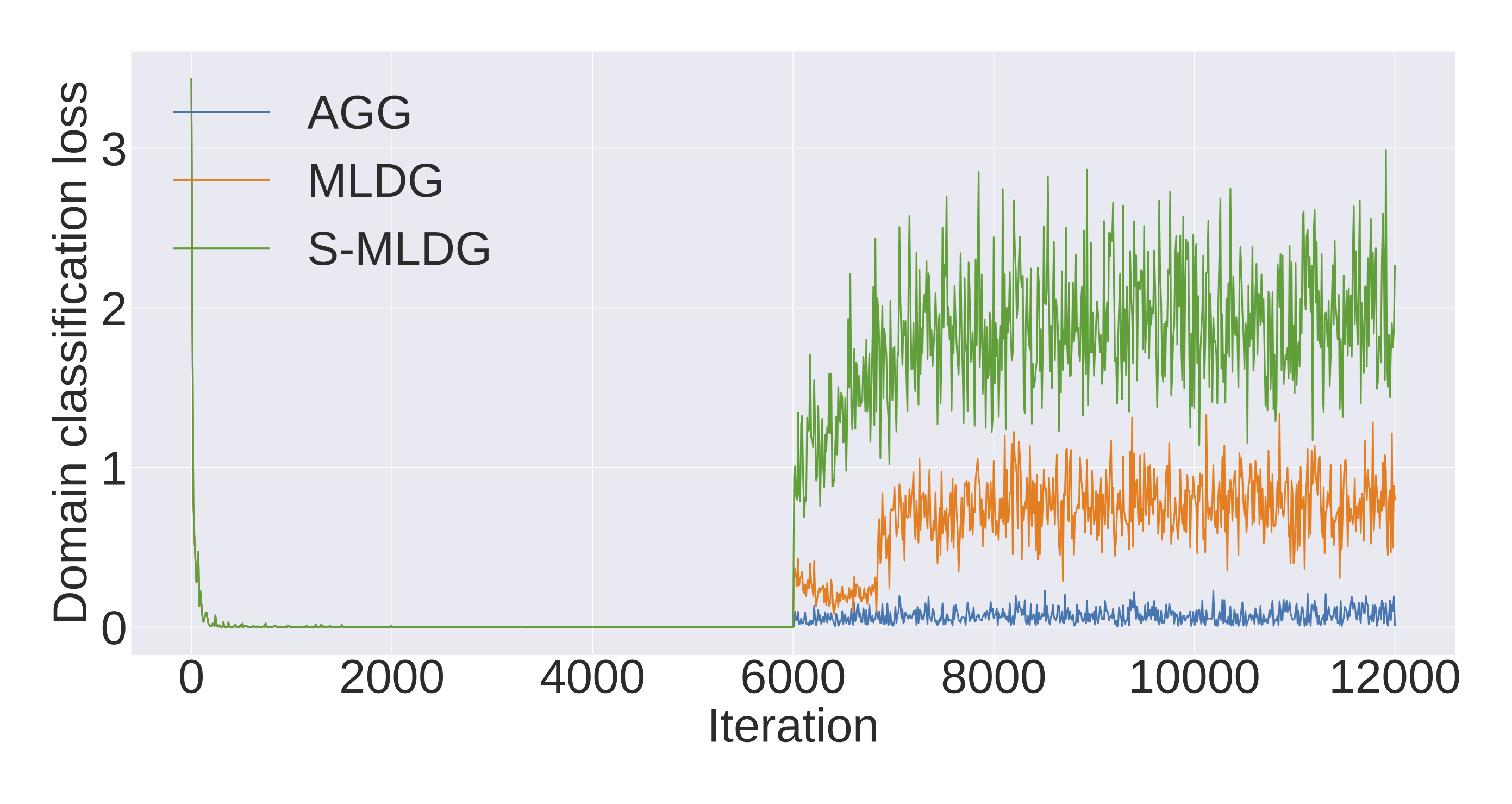}
    % \vspace{-0.3cm}
    \caption{Domain classification loss analysis on VLCS.}
    \label{fig:domain-clf-loss}
\end{figure}

\begin{figure*}[t]
    \centering
    \begin{subfigure}[b]{0.24\linewidth}
    \centering
    \includegraphics[width=1\linewidth]{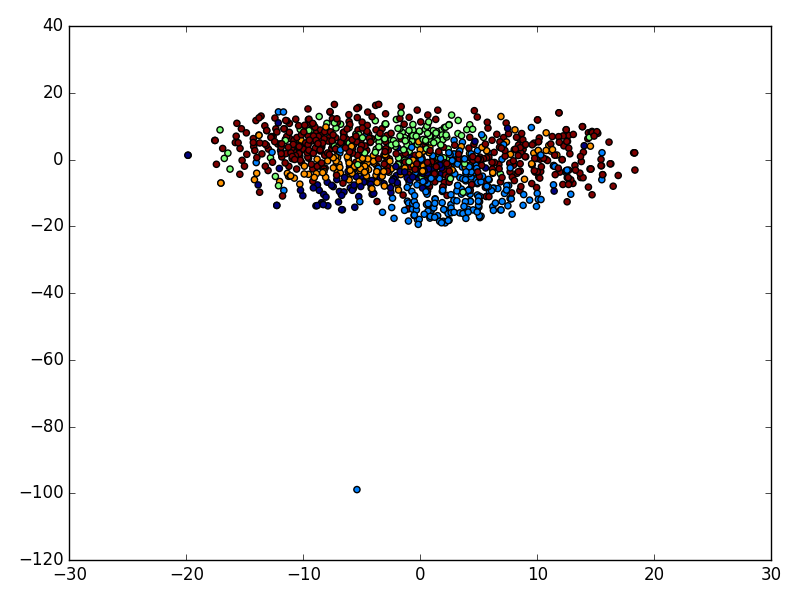}
    % \vspace{-0.7cm}
    \caption{Raw data}
    \end{subfigure}
    \begin{subfigure}[b]{0.24\linewidth}
    \includegraphics[width=1\linewidth]{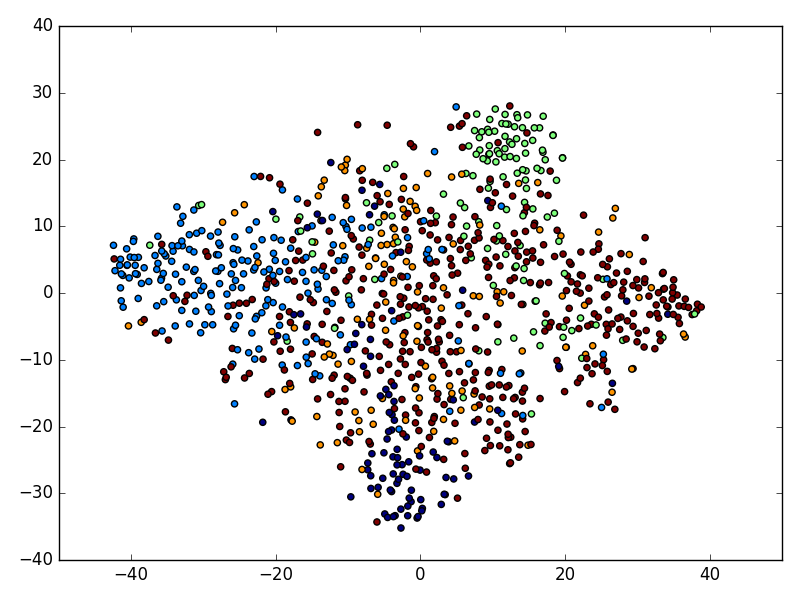}
    %   \vspace{-0.7cm}
    \caption{Naive Ensemble}
     \end{subfigure}
     \begin{subfigure}[b]{0.24\linewidth}
    \includegraphics[width=1\linewidth]{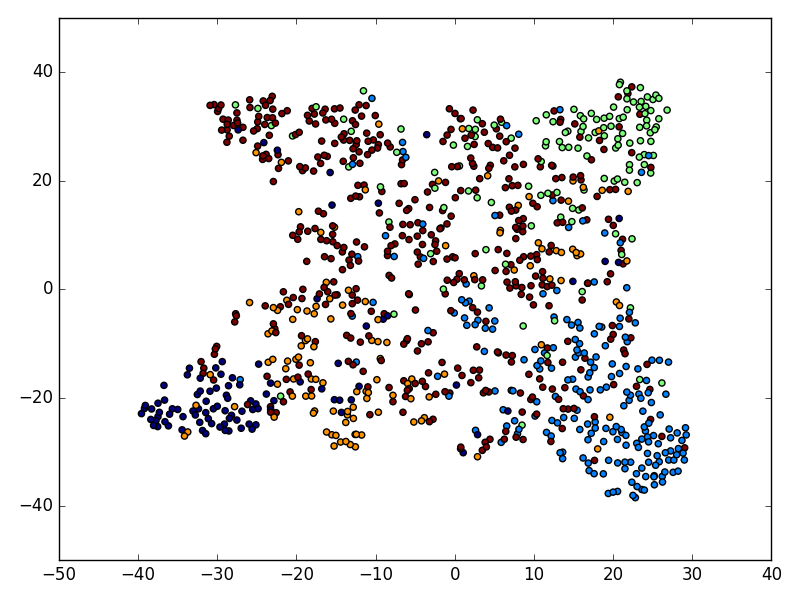}
    %   \vspace{-0.7cm}
    \caption{Undo-Bias}
     \end{subfigure}
     \begin{subfigure}[b]{0.24\linewidth}
    \includegraphics[width=1\linewidth]{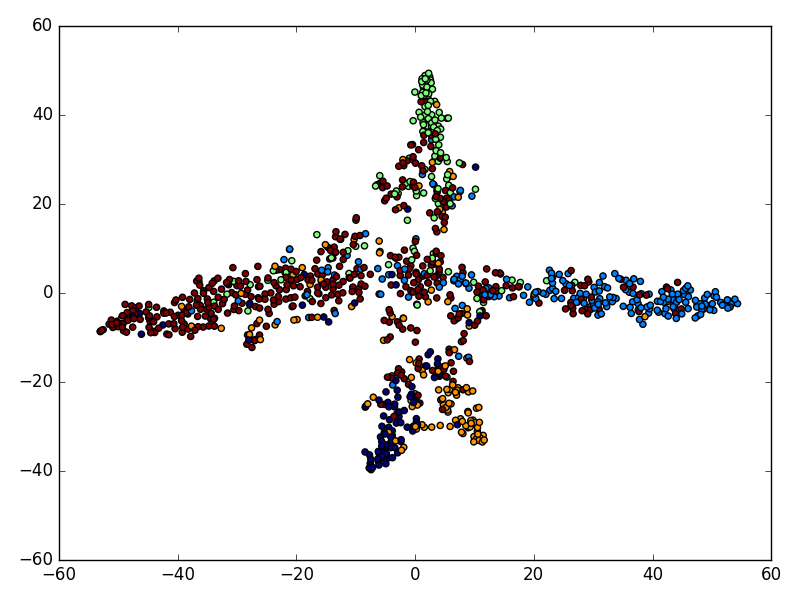}
    %   \vspace{-0.7cm}
    \caption{\hundo{}}
    \end{subfigure}
    % \hfill
    \begin{subfigure}[b]{0.24\linewidth}
    \includegraphics[width=1\linewidth]{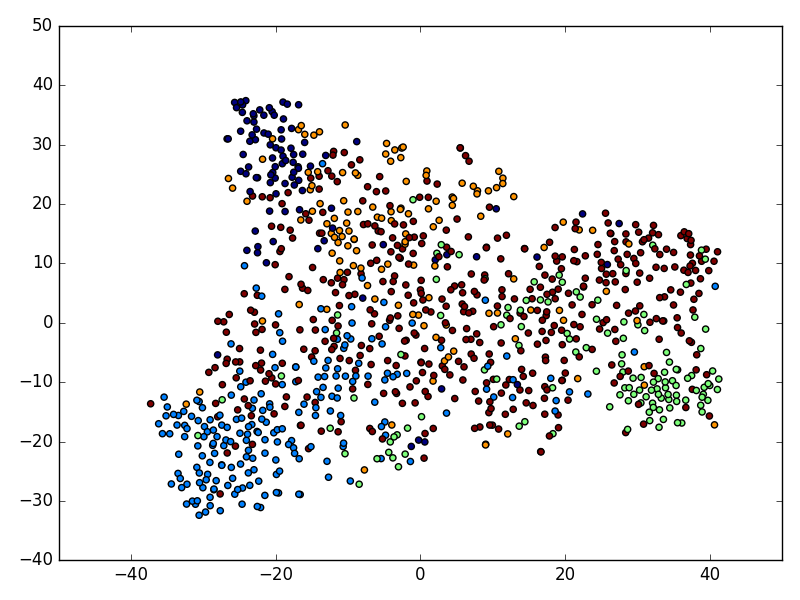}
    %   \vspace{-0.7cm}
    \caption{AGG}
    \end{subfigure}
    \begin{subfigure}[b]{0.24\linewidth}
    \includegraphics[width=1\linewidth]{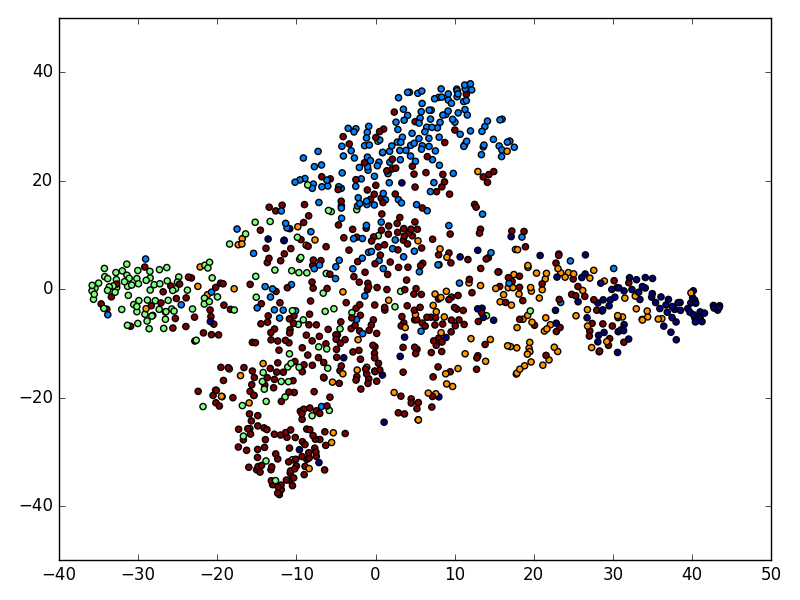}
    %   \vspace{-0.7cm}
    \caption{MLDG}
    \end{subfigure}
    \begin{subfigure}[b]{0.24\linewidth}
     \includegraphics[width=1\linewidth]{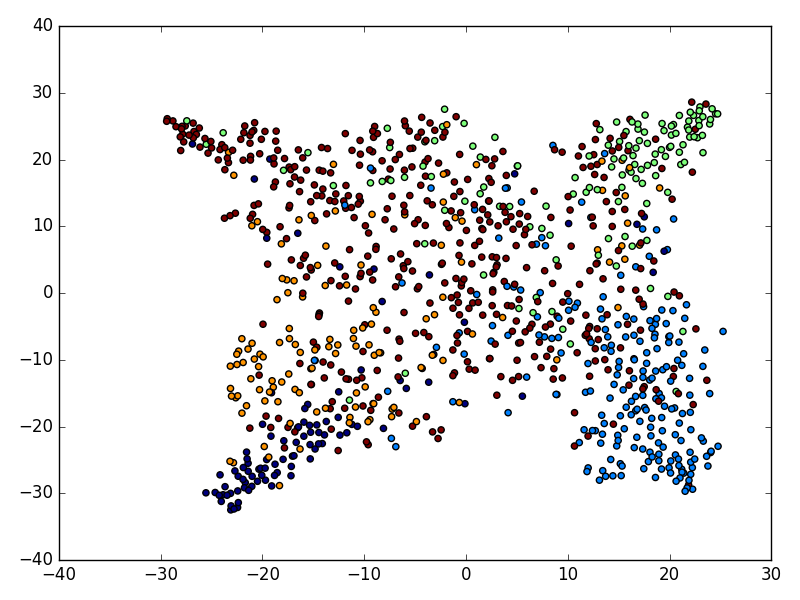}
        % \vspace{-0.7cm}
     \caption{\nameFFS{}}
    \end{subfigure}
    \begin{subfigure}[b]{0.24\linewidth}
    \includegraphics[width=1\linewidth]{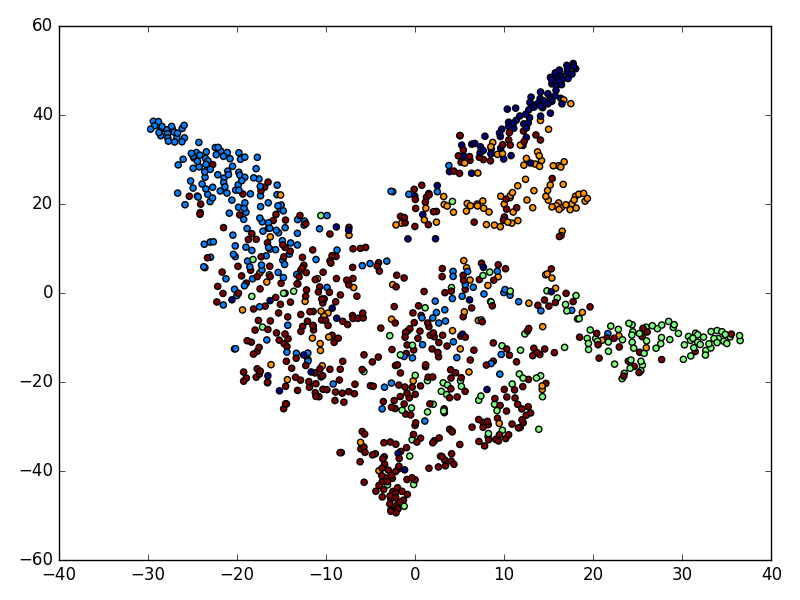}
    %   \vspace{-0.7cm}
    \caption{\nameS{}}
    \end{subfigure}
    % \vspace{-0.2cm}
    \caption{T-SNE visualization of different models' embeddings of VLCS held-out test data (V) after training on (LCS). Colors represent object categories.}
    \label{fig:hmldg-vs-mldg}
\end{figure*}

\keypoint{Results}
%In the results using AlexNet are shown in Table \ref{tab:agg-alex}. We see that, we see that Undo-Bias \cite{ECCV12_Khosla} and MLDG\footnote{We find MLDG has slightly better performance when increasing $\alpha$ and $\beta$ parameters compared to \cite{Li2018MLDG}.} \cite{Li2018MLDG} work well. \hundo again improves over the vanilla Undo-Bias, but do not successfully improve over the vanilla MLDG here, resulting in the shy comparable results. 
From the results in Table~\ref{tab:agg-resnet-pacs}, we can see that: (i) Our sequential learning methods \hundo{} and \nameS{} improve on their counterparts Undo-Bias and MLDG, (ii) \nameFFS{} performs comparably with \nameS{}, and (iii) \nameS{} and \nameFFS{} perform best overall. 
%by upgrading the backbone model, all the methods have got a significant performance boost. By employing this modern architecture, \hundo, FFO-\nameS{} and \nameS{} all have managed to improve the base DG methods. Nevertheless, FFO-\nameS{} and \nameS{} win over all competitors and gain the state-of-the-art results.
% We can see in Table \ref{acc-pacs}, that most evaluated DG methods are better than the simple aggregation (AGG) since there is large domain gap in this benchmark. Among the state-of-the-art baselines, we can see MLDG works well with this large-gap DG -- outperforming DANN, CrossGrad and Undo Bias. Here CrossGrad is not significantly better than the AGG baseline. This highlights a limitation of the recent state of the art CrossGrad \cite{shiv2018dg}. Their assumption is that domains lie on a smooth curve which can be learned and interpolated along using their data augmentation method. Thus while this is suitable for domain shift caused by image rotation or other smooth interpolations, it does not function well for more the challenging shifts between discretely and qualitatively distinct domains like in PACS. In terms of our hierarchical extensions of Undo Bias and MLDG, we see that we achieve a modest improvement on the vanilla versions of each algorithm.

\subsection{Further Analysis}

\keypoint{Analysis for \nameS{}}
As shown earlier, MLDG and \nameS{} aim to maximize the inner-product between gradients of different source domains. Intuitively, optimizing this gradient alignment will lead  to increase domain invariance \cite{ganin2016dann}. To analyze if this is the case, we  use domain-classification loss as a measure of domain invariant feature encoding. We append an additional domain-classifier to the penultimate layer of the original model, creating a domain and category multi-task classifier, where all feature layers are shared.  We train the domain classification task for 6000 iterations, then switch to training the category classification task for another 6000 iterations. Using this setup we compare AGG, MLDG and \nameS{}. From Fig.~\ref{fig:domain-clf-loss}, we see that the domain-classification loss decreases rapidly in the first phase: the domain is easy to recognise before DG training. In the second phase we switch on categorization and DG training. \nameS{} and MLDG give higher domain classification loss than AGG -- indicating that MLDG and \nameS{} learn features that the domain classifier finds harder to distinguish, and hence are the most domain invariant. 
\keypoint{Visualization of Learned Features}
We use t-SNE to visualize the feature embedding of a held-out test domain (V) on VLCS, after training models on L, C and S. From the results in Fig.~\ref{fig:hmldg-vs-mldg}, we can see that before training the raw test data points are not separable by category. As baselines we also compare AGG and Naive Ensemble (training an ensemble of domain-specific models and averaging their result) for comparison to the models of interest: MLDG, \nameS{}, \nameFFS, Undo-Bias and \hundo. We can see that all these DG methods exhibit better separability than the two baselines, with \hundo{} and \nameS{} providing the sharpest separation.

%Although the test data has never been used in the model training, after the model training all the trained models have a sense of the different categories. But more specifically, we can see that the features trained by AGG and naive ensemble do not have a visibly category-separated space for the test data. All the features trained from the advanced methods do the better jobs than these two simple methods. But, we can find our \hundo and \nameS{} provide the most discriminative illustrations on their feature space.

\keypoint{Computational Cost} A major contribution of this paper is a DG strategy that is not only effective but simple (Alg.~\ref{alg:reptile-dg}) and fast to train. To evaluate this we compare the computational cost of of training various methods on PACS with ResNet-18 for $3k$ iterations. We run all the methods on a machine with Intel® Xeon(R) CPU (E5-2687W @ 3.10GHz × 8) and TITAN X (Pascal) GPU. From the results in Table~\ref{tab:computational-cost}, we see that CrossGrad is by far the most expensive with \nameS{} in second place. In contrast, our derived \nameFFS{} is not noticeably slower than the baseline and lower-bound, AGG.  Undo-Bias and \hundo{} run fast due to only applying them into the last layer. But \hundo{} saves training cost over Undo-Bias as explained in \ref{appendix:hundo-bias}. 

\cut{
\begin{table}[t]
    \centering
    \caption{Training cost (mins) for PACS with ResNet-18.}
    \vspace{-0.3cm}
    \label{tab:computational-cost}
    \scalebox{0.52}{
    \begin{tabular}{c|ccccccc}
    \toprule
         & \textbf{AGG} & \textbf{DANN}~\cite{ganin2016dann}  & \textbf{CrossGrad}~\cite{shiv2018dg} & \textbf{MetaReg}~\cite{NIPS2018_metareg} & \textbf{MLDG}~\cite{Li2018MLDG} & \textbf{L-MLDG} & \textbf{\nameFFS{}} \\
         \hline
       Cost  & 10.98 & 11.35& 146.51 &20.01 & 49.77 & 72.64 & 11.04 \\
         \bottomrule
    \end{tabular}
    }
\end{table}}

\begin{table}[t]
    \centering
    \caption{Training cost (mins) for PACS with ResNet-18.}
    % \vspace{-0.2cm}
    \label{tab:computational-cost}
    \resizebox{1.\linewidth}{!}{
    \begin{tabular}{ccccc}
    \toprule
         \textbf{AGG} & \textbf{DANN}~\cite{ganin2016dann}  & \textbf{CrossGrad}~\cite{shiv2018dg} & \textbf{MetaReg}~\cite{NIPS2018_metareg} & \textbf{MLDG} \cite{Li2018MLDG} \\
         \hline
       10.98 & 11.35& 146.51 &20.01 & 49.77 \\
       \hline
        & \textbf{\nameFFS{}}&  \textbf{\nameS{}}  & \textbf{Undo-Bias} \cite{ECCV12_Khosla} & \textbf{\hundo}\\
        \hline
& 11.04& 72.64& 11.16 & 11.01\\
         \bottomrule
    \end{tabular}
    }
\end{table}

\section{Conclusion}
We introduced the idea of sequential learning to provide a training regime for a base DG model. This can be seen as generating more unique DG episodes for learning, and as providing more feedback for back-propagation through the chain of domains. Our framework can be applied to different base DG models including Undo-Bias and MLDG. Our final \nameFFS{} method provides a simple to implement and fast to train DG method that achieves state of the art results on a variety of benchmarks.
%for DG models. This can be seen as increasing the amount of `practice' that a DG model is trained with by considering the exponential number of paths through the available training domains. Our lifelong sequential learning framework is theoretically applicable to all the base DG methods. Empirically, we show that this generalization improves the performance of two different base DG models - Undo Bias and MLDG. By integrating these two base DG methods, they all show the state-of-the-art performances on three different DG benchmarks.

% \clearpage
% \newpage
% \small{
\bibliographystyle{splncs04}
\bibliography{egbib}
% }
\clearpage
\newpage

\appendix

\section{Analysis of Fast First-order \nameS{}\label{app-analysis-ffo}}
If we refer the loss of $i^{th}$ inner-loop step in Alg.~\ref{alg:reptile-dg} as 
\small
\begin{equation}
    \mathcal{L}_{i} =  \mathcal{L}(\mathcal{D}_{p[i]}, \tilde{\theta}_i)
\end{equation}
\normalsize
where $\tilde{\theta}_i$ are the parameters, the gradient of that step is 
\small
\begin{equation}
    g_i=\nabla_{\tilde{\theta}_i}\mathcal{L}_i=\mathcal{L}_i^{'}
\end{equation}
\normalsize
and updated parameters of that step is
\small
\begin{equation}
\tilde{\theta}_{i+1} = \tilde{\theta}_{i} - \alpha g_i
\end{equation}
\normalsize
Then Taylor series of $g_i$ at initial point $\tilde{\theta}_1$ gives 
\small
\begin{equation}
\label{eq:taylor-ser}
\begin{aligned}
g_i =& \mathcal{L}_i^{'}(\tilde{\theta}_1 + \tilde{\theta}_i - \tilde{\theta}_1)\\
=& \mathcal{L}_i^{'}(\tilde{\theta}_1) + \mathcal{L}_i^{''}(\tilde{\phi}_1)(\tilde{\theta}_i - \tilde{\theta}_1) + \mathcal{O}((\tilde{\theta}_i - \tilde{\theta}_1)^2) \\
= & \mathcal{L}_i^{'}(\tilde{\theta}_1) + \mathcal{L}_i^{''}(\tilde{\theta}_1)(\tilde{\theta}_i - \tilde{\theta}_1) + \mathcal{O}(\alpha^2) \\
= & \mathcal{L}_i^{'}(\tilde{\theta}_1) - \mathcal{L}_i^{''}(\tilde{\theta}_1)\sum_{j=1}^{i-1}\alpha g_j + \mathcal{O}(\alpha^2)
\end{aligned}
\end{equation}
\normalsize
where the $\mathcal{O}(\alpha^2)$ items in $g_i$ are omitted due to their small effects in $\alpha g_i$. If we treat the gradient and hessian of $\mathcal{L}_i$ w.r.t $\tilde{\phi}_1$ as $\bar{g}_i$ and $\bar{H}_i$, we have 
\small
\begin{equation}
\begin{aligned}
\bar{g}_i &= \frac{\partial \mathcal{L}_i}{\partial \tilde{\theta}_1}\\ & = \frac{\partial \mathcal{L}_i}{\partial \tilde{\theta}_i} \frac{\partial \tilde{\theta}_i}{\partial \tilde{\theta}_1} \\
          &= g_i \frac{\partial (\tilde{\theta}_1 - \sum_{j=1}^{i-1}\alpha g_j)}{\partial \tilde{\theta}_1} \\
          &= g_i - \mathcal{O}(\alpha)
\end{aligned}
\end{equation}
\normalsize
Equivalently, we get $g_i = \bar{g}_i + \mathcal{O}(\alpha)$. Then together with $\bar{H}_i$, Eq.~\ref{eq:taylor-ser} becomes

\small
\begin{equation}
\label{eq:taylor-ser-init}
\begin{aligned}
g_i &= \bar{g}_i - \bar{H}_i (\sum_{j=1}^{i-1}\alpha (\bar{g}_j + \mathcal{O}(\alpha))) + \mathcal{O}(\alpha^2) \\
&=\bar{g}_i - \alpha \bar{H}_i \sum_{j=1}^{i-1} \bar{g}_j + \mathcal{O}(\alpha^2)
\end{aligned}
\end{equation}
\normalsize
If we consider an example with two source domains $\mathcal{D}_1, \mathcal{D}_2$. We run FFO-\nameS{} with initial parameters $\tilde{\theta}_1$ on $\mathcal{D}_1, \mathcal{D}_2$ recursively, we get two inner-loop steps
\small
\begin{equation}
\label{eq:reptile-grad}
\begin{aligned}
\mathcal{L}_1 = \mathcal{L}(\mathcal{D}_{1}, \tilde{\theta}_1), \quad g_1 = \nabla_{\tilde{\theta}_1}\mathcal{L}_1, \quad \tilde{\theta}_2 = \tilde{\theta}_1 - \alpha g_1 \\
\mathcal{L}_2 = \mathcal{L}(\mathcal{D}_{2}, \tilde{\theta}_2), \quad g_2 = \nabla_{\tilde{\theta}_2}\mathcal{L}_2, \quad \tilde{\theta}_3 = \tilde{\theta}_2 - \alpha g_2
\end{aligned}
\end{equation}
\normalsize
after that we get one-step outer-loop gradient,
\small
\begin{equation}
\label{eq:reptile-grad}
\begin{aligned}
\tilde{\theta}_1 - \tilde{\theta}_3 & = \tilde{\theta}_1 - \tilde{\theta}_2 + \alpha g_2 \\
&= \tilde{\theta}_1 - \tilde{\theta}_1 + \alpha g_1 + \alpha g_2 \\
&= \alpha (g_1 + g_2 )
\end{aligned}
\end{equation}
\normalsize
And when we bring Eq.~\ref{eq:taylor-ser-init} in, we get
\small
\begin{equation}
\begin{aligned}
g_1 + g_2
&= \bar{g}_1 + \bar{g}_2 - \alpha \bar{H}_2 \bar{g}_1 + \mathcal{O}(\alpha^2)
\end{aligned}
\end{equation}
\normalsize
then, if we shuffle the order of $\mathcal{D}_1$, $\mathcal{D}_2$ and run on $\mathcal{D}_2$, $\mathcal{D}_1$ recursively, we get
\small
\begin{equation}
\begin{aligned}
g_2 + g_1
&= \bar{g}_2 + \bar{g}_1 - \alpha \bar{H}_1 \bar{g}_2 + \mathcal{O}(\alpha^2)
\end{aligned}
\end{equation}
\normalsize

\noindent Taking the expectation over these two sequences, we get
\small
\begin{equation}
\label{eq:ffo-hmldg-expectation-two-loss}
\begin{aligned}
 \mathbb{E}_{p\sim\mathcal{P}}[g_{p[1]}+g_{p[2]}] = \bar{g}_1 + \bar{g}_2 + \frac{1}{2} (- \alpha \bar{H}_1 \bar{g}_2 - \alpha \bar{H}_2 \bar{g}_1 ) + \mathcal{O}(\alpha^2)\\
\end{aligned}
\end{equation}
\normalsize
The first term $\bar{g}_1 + \bar{g}_2$ in Eq.~\ref{eq:ffo-hmldg-expectation-two-loss} is the gradient that minimizes the losses on $\mathcal{D}_1$, $\mathcal{D}_2$. The second term is
\small
\begin{equation}
    \begin{aligned}
    \frac{1}{2} (- \alpha \bar{H}_1 \bar{g}_2 - \alpha \bar{H}_2 \bar{g}_1 ) = - \frac{\alpha}{2} \frac{\partial (\bar{g}_1 \cdot \bar{g}_2)}{\partial \tilde{\theta}_1}
    \end{aligned}
\end{equation}
\normalsize
% Here $\bar{g}_1 \cdot \bar{g}_2$ is the inner product between the two gradients. The gradient $ - \frac{\partial (\bar{g}_1 \cdot \bar{g}_2)}{\partial \tilde{\theta}_1}$ is in the direction that maximizes it. This means in expectation of multiple gradient updates \nameFFS{} learns to maximize the inner-product between gradients of different domains. Thus it maintains a similar but slightly different objective to \nameS{}, which maximizes the inner-product of gradients in each meta update.

\section{Application to Undo Bias}\label{appendix:sec:undo}
\subsection{Reinterpreting Vanilla Undo Bias}
\keypoint{Background}
\emph{Undo Bias} is a classic domain generalization method that was initially proposed specifically for DG with shallow linear classifiers \cite{ECCV12_Khosla}, although it has been extended to the multi-linear setting for end-to-end deep learning \cite{da2017dg}. The hypothesis is that classifiers for all domains (datasets) can be modelled as the sum of an underlying domain-agnostic model $\theta_0$ and a domain-specific bias $\vec{\theta}_i$ for each domain $i$. With this assumption, the objective for training on all $N$ source domains in $\mathcal{D}$ is,
\small
\begin{equation}
\label{eq:Undo-Bias}
\underset{\vec{\theta}_{0}, \vec{\theta}_{1},\dots \vec{\theta}_{N}}{\operatorname{argmin}}~ \sum_{i=1}^{N} \| \mathit{X}_{i}(\vec{\theta}_{0} + \mathit{\vec{\theta}_{i}}) - y_{i}  \|_{2}^{2} +\lambda_{1} \sum_{i=1}^{N} \|\mathit{\vec{\theta}_{i}} \|_{2}^{2} + \lambda_{2} \|\vec{\theta}_{0} \|_{2}^{2} 
\end{equation}
\normalsize
where $\lambda_{1}$ and $\lambda_{2}$ are regularizer weights.
After training, the shared parameter $\theta_0$ is assumed to represent a domain-agnostic classifier and used for inference on unseen domains.

\keypoint{Reinterpretation}
We can deduce an equivalent formula to Eq.~\ref{eq:Undo-Bias} expressed only in terms of domain-specific models $\vec{\theta}_i$
\small
\begin{equation}
\label{eq:Undo-Bias-deduce-r}
\begin{aligned}
\underset{\vec{\theta}_{1},\dots \vec{\theta}_{N}}{\operatorname{argmin}}~ &\sum_{i=1}^{N} \| \mathit{X}_{i}\vec{\theta}_{i} - y_{i}  \|_{2}^{2} + \frac{\lambda_{1}\lambda_{2}}{\lambda_{2} + \lambda_{1}N} \sum_{i=1}^{N} \|\vec{\theta}_{i} \|_{2}^{2} \\ 
& + \frac{{\lambda_{1}^{2}N}}{\lambda_{2} + \lambda_{1}N} \sum_{i=1}^{N} \| \vec{\theta}_{i} - \frac{\sum_{j=1}^{N}\vec{\theta}_{j}}{N}   \|_{2}^{2}
\end{aligned}
\end{equation}
\normalsize
In this equivalent case, the model parameter to use for unseen domains is $ \frac{\sum_{i=1}^{N}\vec{\theta}_{i}}{N}$ (i.e., the underlying domain should be close to the mean of the parameters of all source domains). This alternative formulation will be useful for the hierarchical extension later. While the presentation so far is for a regression problem with MSE loss, the general form of Eq.~\ref{eq:Undo-Bias-deduce-r} for any loss function $\mathcal{L}(\cdot)$ can be written as,
\small
\begin{equation}
\label{eq:Undo-Bias-deduce}
\underset{\vec{\theta}_{1},\dots \vec{\theta}_{N}}{\operatorname{argmin}}~ \sum_{i=1}^{N} \mathcal{L}(\mathcal{D}_i, {\theta}_{i}) + \lambda \|\vec{\theta}_{i} - \frac{\sum_{j=1}^{N}\vec{\theta}_{j}}{N} \|_{2}^{2}
\end{equation}
\normalsize
Here we omit the second term in Eq.~\ref{eq:Undo-Bias-deduce-r}, i.e., the squared $\ell_2$ norm on parameter, because it is usually realised by weight decay when training a neural network model. And empirically, we find using $\|.\|_{2}$ for the second item in Eq.~\ref{eq:Undo-Bias-deduce} is easier to tune.

\keypoint{Derivation for Eq.~\ref{eq:Undo-Bias-deduce-r} \label{appendix-deduction}}
\begin{equation}
\label{supp:eq:Undo-Bias}
\underset{\vec{\theta}_{0}, \theta_{1},\dots \theta_{N}}{\operatorname{argmin}}~ \sum_{i=1}^{N} \| \mathit{X}_{i}(\vec{\theta}_{0} + \mathit{\vec{\theta}_{i}}) - y_{i}  \|_{2}^{2} +\lambda_{1} \sum_{i=1}^{N} \|\mathit{\vec{\theta}_{i}} \|_{2}^{2} + \lambda_{2} \|\vec{\theta}_{0} \|_{2}^{2} 
\end{equation}
We denote the objective function in Eq.~\ref{supp:eq:Undo-Bias} as $\mathcal{L}(\theta, \theta_0)$, where we regard the optimal solution for Eq.~\ref{supp:eq:Undo-Bias} is $\theta^*$ and $\theta_0^*$, then we have $\frac{\partial \mathcal{L}}{\partial \theta_{i}} |_{\theta_{i}=\theta_i^*, \theta_0 = \theta_0^{*}} = 0, \forall i \in [1,2,\dots,N]$ and $\frac{\partial \mathcal{L}}{\partial \theta_{0}} |_{\theta_{i}=\theta_i^*, \theta_0 = \theta_0^{*}} = 0$.

Given this solution, we have,
\begin{equation}
{X_{i}}^T (X_{i}\theta_{i}^* + X_{i}\theta_{0}^* - y^{(i)}) + \lambda_1 \theta_{i}^* = 0
\label{supp:eq:mtlam4}
\end{equation}
and
\begin{equation}
\sum_{i=1}^{T} {X_{i}}^T (X_{i}\theta_{i}^* + X_{i}\theta_{0}^* - y^{(i)}) + \lambda_2 \theta_{0}^* = 0
\label{supp:eq:mtlam5}
\end{equation}
When we aggregate all Eq.~\ref{supp:eq:mtlam4}, we get,
\begin{equation}
\sum_{i=1}^{N}{X_{i}}^T (X_{i}\theta_{i}^* + X_{i}\theta_{0}^* - y^{(i)}) + \lambda_1 \sum_{i=1}^{N}\theta_{i}^* = 0
\label{supp:eq:mtlam6}
\end{equation}
After the subtraction of the common items in the Eq.~\ref{supp:eq:mtlam5} and Eq.~\ref{supp:eq:mtlam6}, we get,
\begin{equation}
\theta_0^{*} = \frac{\lambda_1}{\lambda_2} \sum_{i=1}^N \theta_i^*
\label{supp:eq:mtlam7}
\end{equation}
If we assume for each specific domain $i$, the parameterized weights are $\Theta_i= \theta_i + \theta_0$, then we combine this with Eq.~\ref{supp:eq:mtlam7}, and further obtain that,
\begin{equation}
\theta_0^{*} = \frac{\lambda_1}{\lambda_2 + \lambda_1 N} \sum_{i=1}^N \Theta_i^* = \frac{1}{\frac{\lambda_2}{\lambda_1} + N} \sum_{i=1}^{N} \Theta_i^*
\label{supp:eq:mtlam8}
\end{equation}
The obtained Eq.~\ref{supp:eq:mtlam8} indicates that the shared parameters $\theta_0$ is a (slightly smoothed) average of all the domain-specific parameters $\Theta_i$. Therefore, we can get 
\begin{equation}
\label{supp:eq:Undo-Bias-deduce-r}
\begin{aligned}
\underset{\Theta_{1},\dots \Theta_{N}}{\operatorname{argmin}}~ & \sum_{i=1}^{N} \| \mathit{X}_{i}\Theta_{i} - y_{i}  \|_{2}^{2} 
+ \frac{\lambda_{1}\lambda_{2}}{\lambda_{2} + \lambda_{1}N} \sum_{i=1}^{N} \|\Theta_{i} \|_{2}^{2} \\ 
& + \frac{{\lambda_{1}^{2}N}}{\lambda_{2} + \lambda_{1}N} \sum_{i=1}^{N} \| \Theta_{i} - \frac{\sum_{j=1}^{N}\Theta_{j}}{N}   \|_{2}^{2}
\end{aligned}
\end{equation}

\subsection{Sequential Undo Bias}\label{appendix:hundo-bias}

\begin{algorithm}[t]
\SetAlgoLined
\textbf{Input}:$\mathcal{D} = [\mathcal{D}_1, \mathcal{D}_2, \dots, \mathcal{D}_N]$\\
\textbf{Initialize}: $\lambda$, $\gamma$ and $[\theta_1,\theta_2,\dots,\theta_N]$\\
 \While{not done training}{
  $p=\operatorname{shuffle}([1,2,\dots,N])$~~~//\small{Randomly sample a trajectory} \\
  $\tilde{\mathcal{D}} = [\tilde{\mathcal{D}}_1, \tilde{\mathcal{D}}_2, \dots, \tilde{\mathcal{D}}_N]$~~~//\small{Sample a mini-batch $\tilde{\mathcal{D}}_i$ for each domain $\mathcal{D}_i$}\\
  $\mathcal{L} = \mathcal{L}(\tilde{\mathcal{D}}_{p[1]}, \theta_{p[1]})$ \\
  \For{$i$ \textbf{in} $[2,3,\dots,|p|]$}
  {
  $\mathcal{L}$ += $\Big(\mathcal{L}(\tilde{\mathcal{D}}_{p[i]}, \theta_{p[i]}) + \lambda \| \theta_{p[i]} - \frac{\sum_{j=1}^{i-1}\theta_{p[j]}}{i-1} \|_2^2 \Big)$\\
  }
  //\small{One-step \hundo{} update}\\
  Update $\theta_i := \theta_i -  \gamma\nabla_{\theta_i}\mathcal{L} $~~~
  
 }
 \textbf{Output}: $\frac{\sum_{i=1}^{N} \theta_{1},\theta_{2},\dots,\theta_{N}}{N}$
 \caption{\hundo: Sequential Undo Bias}
 \label{alg:hundobias}
\end{algorithm}

Vanilla Undo Bias aims to learn an underlying domain-agnostic model with one optimization on a fixed set of source domains (Section~\ref{appendix:sec:undo}). To instantiate our proposed framework (Eq.~\ref{eq:h}) for Undo Bias, we need to extend it to sequential learning. Intuitively, for a given sequence of training domains, we should learn an underlying domain from the first two, and then update this when the third domain comes in, etc. Building on the Undo-Bias formulation in Eq.~\ref{eq:Undo-Bias-deduce}, we define the objective: 
\small
\begin{equation}
\label{eq:h-udl}
\begin{aligned}
\mathcal{L}_{\text{\hundo{}}} = & \mathbb{E}_{p\sim\mathcal{P}}~ \mathcal{L}(\mathcal{D}_{p[1]}, \theta_{p[1]}) \\ 
& + \sum_{i=2}^{N} \Big( \mathcal{L}(\mathcal{D}_{p[i]}, \theta_{p[i]}) + \lambda \| \theta_{p[i]} - \bar{\theta}_{i}) \|_2^2 \Big) 
\end{aligned}
\end{equation}
\normalsize

\noindent where $p$ is a path through all possible permutations of domains $\mathcal{P}$, and $i\in p$ iterates over that path. $\bar{\theta}_{i}$ is the running average over the parameters in the path \emph{before} it arrives at $\theta_{p[i]}$, i.e., $\bar{\theta}_{i}=\frac{\sum_{j=1}^{i-1}\theta_{p[j]}}{i-1}$. The first term is not directly path-dependent, but it becomes so via shared parameters with the second path-dependent term. In this objective, when training $\theta_i$ for domain $i$, backpropagation also updates all domains in the path before domain $i$. We term this procedure Sequential Undo Bias. And its algorithm flow is shown in Alg.~\ref{alg:hundobias}.

To unpack Eq.~\ref{eq:h-udl}, we use a length-3 path example. The objective function is then:
\small
\begin{equation}
\label{eq:h-udl-example}
\begin{aligned}
\mathcal{L}_{\text{\hundo{}-3}} = \mathbb{E}_{p\sim\mathcal{P}}~~~ & \mathcal{L}(\mathcal{D}_{p[1]}, \theta_{p[1]}) \\
+& \mathcal{L}(\mathcal{D}_{p[2]}, \theta_{p[2]}) + \lambda \| \theta_{p[2]} - \theta_{p[1]} \|_2^2  \\
+& \mathcal{L}(\mathcal{D}_{p[3]}, \theta_{p[3]}) + \lambda \| \theta_{p[3]} - \frac{\theta_{p[1]}+\theta_{p[2]}}{2} \|_2^2
\end{aligned}
\end{equation}\normalsize
This says: train vanilla Undo Bias on the first two domains (first three terms), and then incrementally train Undo Bias for the third domain (fourth and fifth term). If the first Undo Bias model $\frac{\theta_{p[1]}+\theta_{p[2]}}{2}$ was fixed after training $\theta_{p[1]}$ and $\theta_{p[2]}$ then this would be simple regularization of $\theta_{p[3]}$ training by the Undo Bias source (fifth term regularizer). But backpropagating means that $\theta_{p[1]}$ and $\theta_{p[2]}$ are trained so as not only to solve their domains in an Undo Bias way, but also to help learn $\theta_{p[3]}$. This is a DG `practice' for the first trained domains-specific parameters. Finally, the optimization should be applied for all possible permutations of $[1,2,3]$. In this example $\frac{\theta_1 + \theta_2 + \theta_3}{3}$ would then be used as the final Undo Bias model for the true testing domain. And as the sequential path goes deeper, the former ranking domains get more `practices'.

\keypoint{Computational Cost}
The difference between Undo-Bias and \hundo{} can be found by comparing Eq.~\ref{eq:Undo-Bias-deduce} and Eq.~\ref{eq:h-udl}. We can see the computational difference happens in the second terms in the objective functions. In vanilla Undo-Bias, each domain-specific parameter would have a L2 loss to minimize its difference to the mean of all domain-specific parameters. But in \hundo, due to the hierarchical structure, each domain would have the same L2 loss to the parameters of the traversed domains. If we regard the computational complexity of the L2 loss of $n$ domain-specific parameters as $\mathcal{O}(n)$, the computational complexity for the second item of \hundo{} is $\sum_{i=2}^{n-1} \mathcal{O}(i)$, which is smaller than that of Undo-Bias $ n\mathcal{O}(n)$. So, due to the hierarchical learning, \hundo{} saves computation over Undo-Bias. This is proved in the training cost comparison in Table~\ref{tab:computational-cost}.

\begin{figure}[t]
    \centering
    \centering
    \includegraphics[width=1\linewidth]{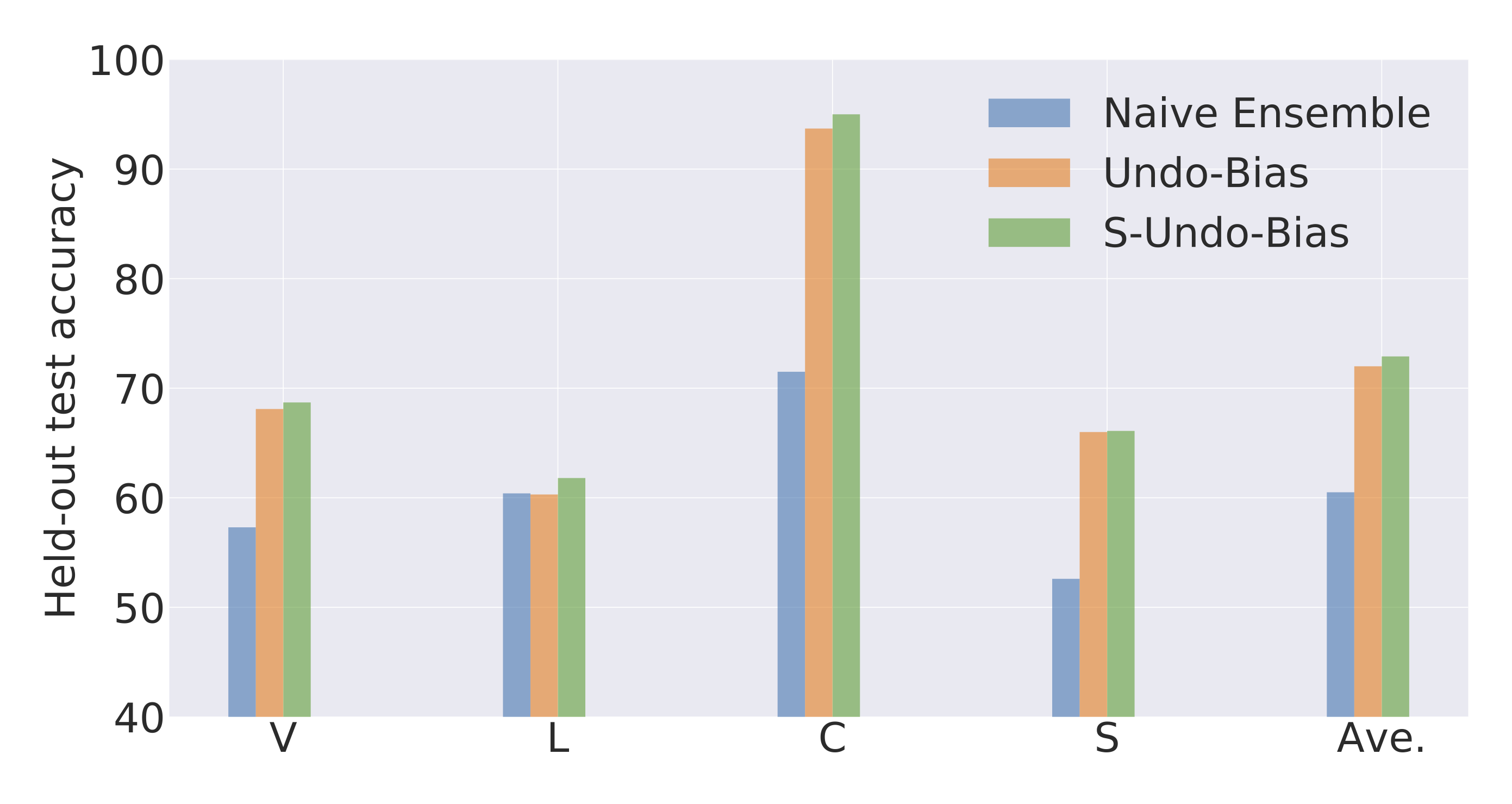}
    % \vspace{-0.7cm}
    \caption{Validation of reformulated (S)-Undo-Bias.}
    \label{fig:hundo-vs-naive}
\end{figure}

\keypoint{Validation of Reformulated (S)-Undo-Bias} We validate our reformulated Undo-Bias and \hundo{} on VLCS by comparison to a naive ensemble, which trains all the domain specific branches separately and uses fused ensemble of models at inference. The comparison in Fig.~\ref{fig:hundo-vs-naive} shows that both Undo-Bias and \hundo{} learn better solutions than the simple ensemble of domain-specific models. This shows that the performance gains of our reformulated Undo-Bias and \hundo{} are not merely due to the model ensemble.

\section{Training Hyper Parameters}

We can set different $\alpha$ and $\beta$ (=1 by default) for different inner loops in Alg.~\ref{alg:full} and \ref{alg:reptile-dg} and refer $\alpha_i$ and $\beta_i$ as the coefficients in the $i^{\text{th}}$ inner loop.
We use M-SGD with momentum=0.9, weight decay=0.00005.

\keypoint{IXMAS} 
\begin{itemize}
    \item \emph{\nameS{}}: $\alpha_1$=$\alpha_2$=$\alpha_3$=0.9, $\gamma$=0.001 and $\beta_{4}$=2.0.
    \item \emph{\nameFFS{}}: $\alpha_1$=$\alpha_2$=$\alpha_3$=1.0, $\gamma$=0.9 and $\beta_{4}$=1.1.
    \item \emph{\hundo}: $\gamma$=0.005 and $\lambda$=1000.0.
\end{itemize}

\keypoint{VLCS} 
\begin{itemize}
    \item \emph{\nameS{}}: $\alpha_1$=0.05, $\alpha_2$=0.6, $\gamma$=0.001 and $\beta_{3}$=1.2.
    
    \item \emph{\nameFFS{}}: $\alpha_1$=$\alpha_2$=0.3, $\gamma$=0.01 and $\beta_{3}$=1.5.
    
    \item \emph{\hundo}: $\gamma$=0.01 and $\lambda$=50.0.
\end{itemize}

\cut{
\keypoint{PACS(A)} 
\begin{itemize}
    \item \emph{\nameS{}}: $\alpha_1$=$\alpha_2$=0.001, $\gamma$=0.0005 and $\beta_{2}$=4.0.
     \item \emph{\nameFFS{}}: $\alpha_1$=$\alpha_2$=0.005, $\gamma$=0.85 and $\beta_{2}$=1.85.
    \item \emph{\hundo}: $\gamma$=0.001 and $\lambda$=300.0.
\end{itemize}}

\keypoint{PACS} 

\begin{itemize}
    \item \emph{\nameS{}}: $\alpha_1$=$\alpha_2$=0.002, $\gamma$=0.001 and $\beta_{3}$=1.85.
    \item \emph{\nameFFS{}}: $\alpha_1$=$\alpha_2$=0.01, $\gamma$=0.9 and $\beta_{3}$=1.75.
    \item \emph{\hundo}: $\gamma$=0.001 and $\lambda$=100.0.
\end{itemize}

\end{document}